%% file: ms.tex
\pdfoutput=1

\documentclass[journal,transmag]{IEEEtran}
\ifCLASSINFOpdf
\usepackage[pdftex]{graphicx}
\else
\fi

\usepackage{amssymb,amsmath}
\usepackage{latexsym}
\usepackage{url}
\usepackage{xcolor}
\definecolor{newcolor}{rgb}{.8,.349,.1}
\usepackage{algorithm}
\usepackage[noend]{algpseudocode}
\usepackage{bbm}
\usepackage{dsfont}
\usepackage{adjustbox}
\usepackage{amssymb}
\usepackage{latexsym}
\usepackage{url}
\usepackage{xcolor}
\usepackage{amssymb,amsmath}
\usepackage{latexsym}
\usepackage{url}

\def\eg{\emph{e.g.}}

\renewcommand{\Re}[1]{\mbox{$\mathbbm{R}^{#1}$}}

\newcommand{\tuple}[1]{\mbox{$\mathbbm{#1}$}}

\def\indi#1{\mbox{$\mathds{1}_{#1}$}}

\begin{document}
	\title{A Multiple Source Hourglass Deep Network \\ for Multi-Focus Image Fusion}
	
	%,~\IEEEmembership{Student,~IEEE}
	\author{\IEEEauthorblockN{Fidel A. Guerrero Pe\~{n}a\IEEEauthorrefmark{1,2},
			Pedro D. Marrero Fern\'{a}ndez\IEEEauthorrefmark{1},
			Tsang Ing Ren\IEEEauthorrefmark{1}, \\
			Germano C. Vasconcelos\IEEEauthorrefmark{1}, and
			Alexandre Cunha\IEEEauthorrefmark{2}}
		
		\IEEEauthorblockA{\IEEEauthorrefmark{1}Centro de Inform\'{a}tica (CIn), Universidade Federal de Pernambuco (UFPE), Brazil}
		\IEEEauthorblockA{\IEEEauthorrefmark{2}Center for Advanced Methods in Biological Image Analysis (CAMBIA), California Institute of Technology, USA}% <-this % stops an unwanted space
		\thanks{
			Corresponding author: Fidel Guerrero Pe\~{n}a (email: fagp@cin.ufpe.br).}}
	
	\markboth{Submitted to IEEE Transactions on Image Processing}%
	{Shell \MakeLowercase{\textit{et al.}}: Bare Demo of IEEEtran.cls for IEEE Transactions on Image Processing}

	\IEEEtitleabstractindextext{%
		\begin{abstract}
			Multi-Focus Image Fusion seeks to improve the quality of an acquired burst of images with different focus planes. For solving the task, an activity level measurement and a fusion rule are typically established to select and fuse the most relevant information from the sources. However, the design of this kind of method by hand is really hard and sometimes restricted to solution spaces where the optimal all-in-focus images are not contained. Then, we propose here two fast and straightforward approaches for image fusion based on deep neural networks. Our solution uses a multiple source Hourglass architecture trained in an end-to-end fashion. Models are data-driven and can be easily generalized for other kinds of fusion problems. A segmentation approach is used for recognition of the focus map, while the weighted average rule is used for fusion. We designed a training loss function for our regression-based fusion function, which allows the network to learn both the activity level measurement and the fusion rule. Experimental results show our approach has comparable results to the state-of-the-art methods with a 60X increase of computational efficiency for $520 \times 520$ resolution images. 
		\end{abstract}
		
		\begin{IEEEkeywords}
		Hourglass network, Deep Learning, Multi-Focus Image Fusion.
		\end{IEEEkeywords}}
        \maketitle
		
		\IEEEdisplaynontitleabstractindextext

		\IEEEpeerreviewmaketitle

\input{introduction}
\input{related}
\input{method}
\input{results}
\section{Conclusions}
\label{sec:conclusion}
This paper presented two multi-source hourglass architectures for the multi-focus image fusion problem. The segmentation approach learns the activity level measurement by estimating the focus map of the sources. Then, the weighted average rule is applied to the fusion step. Our regression approach achieved comparable results to state-of-the-art available approaches while trained to learn both the activity level measurement and the fusion rule at once. Experiments with synthetic and real data sets evidenced the feasibility of our methods for two and multiple sources fusion. The main advantages of our approach are its simplicity and considerably improved speed when compared to current multi-focus image fusion methods while maintaining an excellent performance level. The generality of the HF-Reg approach shows the viability to perform other kinds of task like the multi-focus image fusion of noisy inputs.

\section*{Acknowledgment}
The authors thanks the financial support from the Brazilian funding agencies FACEPE, CAPES and CNPq. 
\bibliographystyle{IEEEtran}
\bibliography{refs}
\end{document}

%% file: introduction.tex
\section{Introduction}
% [BEGIN WITH IMAGE FUSION, THEN CONTINUE WITH MFIF AND APPLICATIONS. PROBLEMS IN MFIF METHODS (SIAMESES ARCHITECTURES, PATCHES, SSIM IS NOT THE BEST, LIMITED TO TWO IMAGES). OUR PROPOSAL/CONTRIBUTIONS]

Usually, the limited depth-of-field operation of digital cameras causes only one plane image to stay in focus while the others appear blurred. This focus plane is composed of all objects near a fixed focus point. Taking several shots with different focus points allows the capture of a burst of images where all focus planes become available. The process of reconstructing the entirely focused image by estimating the sharpest pixel values using frame information is named Multi-Focus Image Fusion (MFIF). The resulting focused image is known in the literature as the all-in-focus image and is typically used for further computer processing. Thus, MFIF can be described as a pre-processing step that improves the quality of the acquired burst of images \cite{tan2017automated,tang2018pixel}. Applications of MFIF include, but are not limited to, medical and biological imaging, video surveillance and digital photography \cite{gangapure2015steerable, kong2014adaptive}. Many challenges, such as identifying the focus map in each frame, selecting the fusion function to combine the focus planes and performing a quick and reliable combination of images, remain as open issues, making the multi-focus image fusion an interesting problem to investigate.

\begin{figure}[t!]
\footnotesize
\begin{center}
\centering
\setlength{\tabcolsep}{1pt}
\begin{tabular}{ccc}
\includegraphics[width=0.32\linewidth]{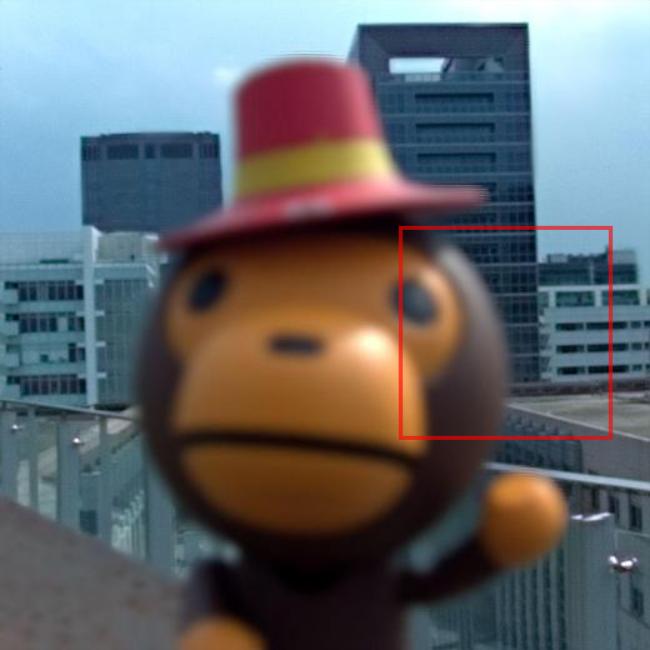}&
\includegraphics[width=0.32\linewidth]{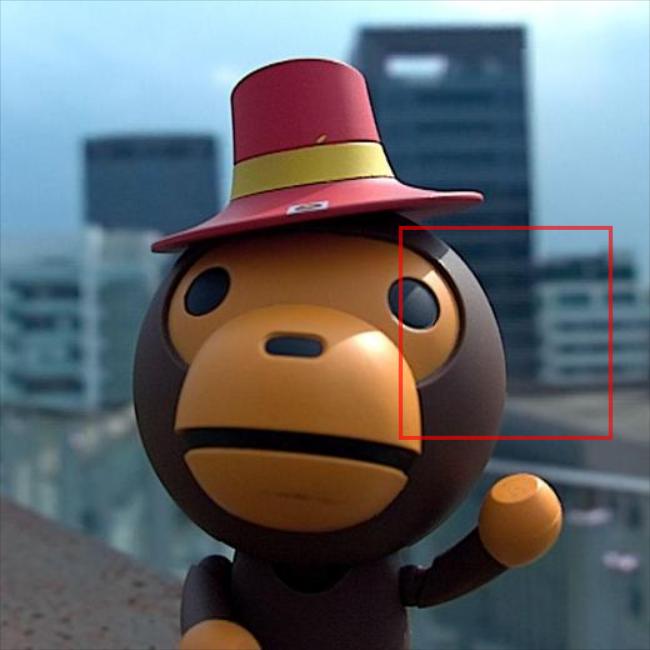}&
\includegraphics[width=0.32\linewidth]{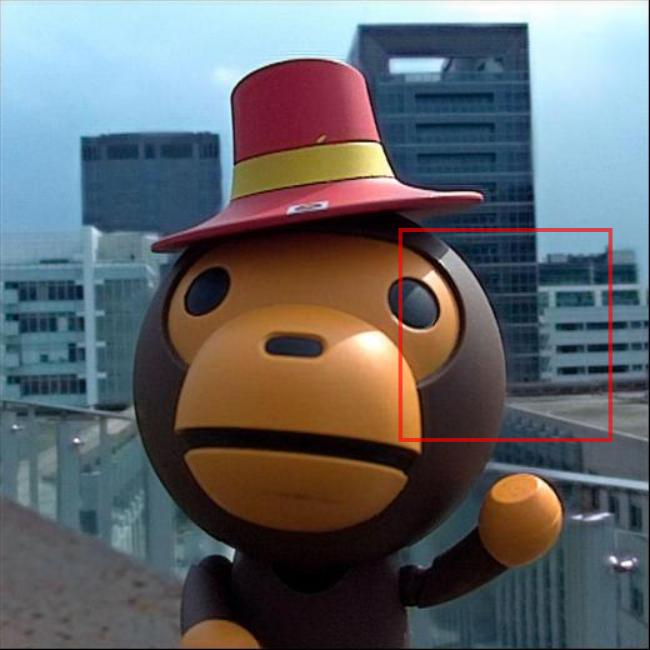}\\
\includegraphics[width=0.32\linewidth]{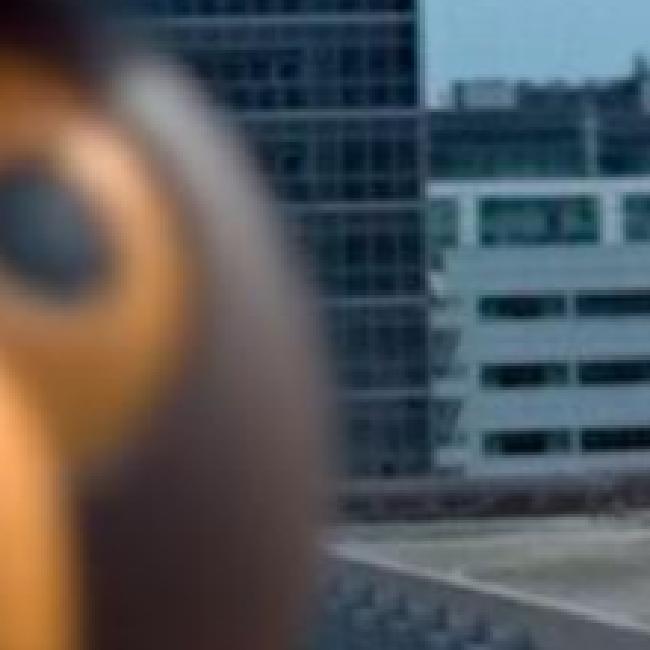}&
\includegraphics[width=0.32\linewidth]{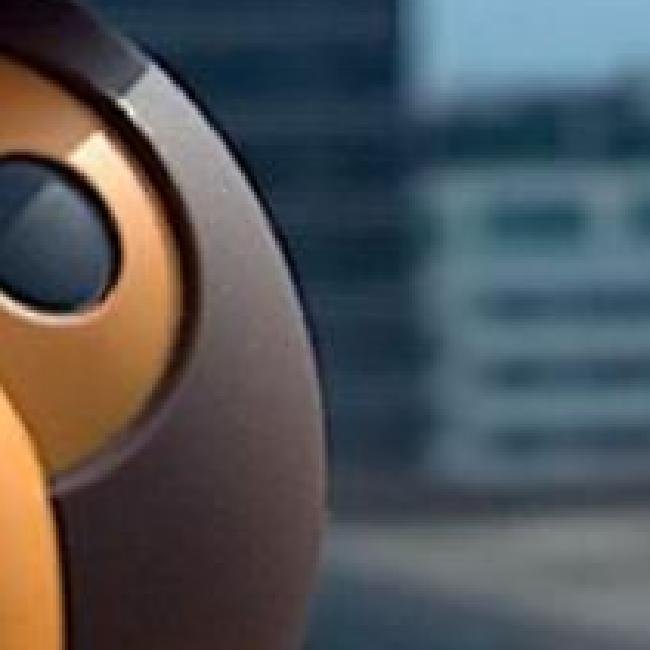}&
\includegraphics[width=0.32\linewidth]{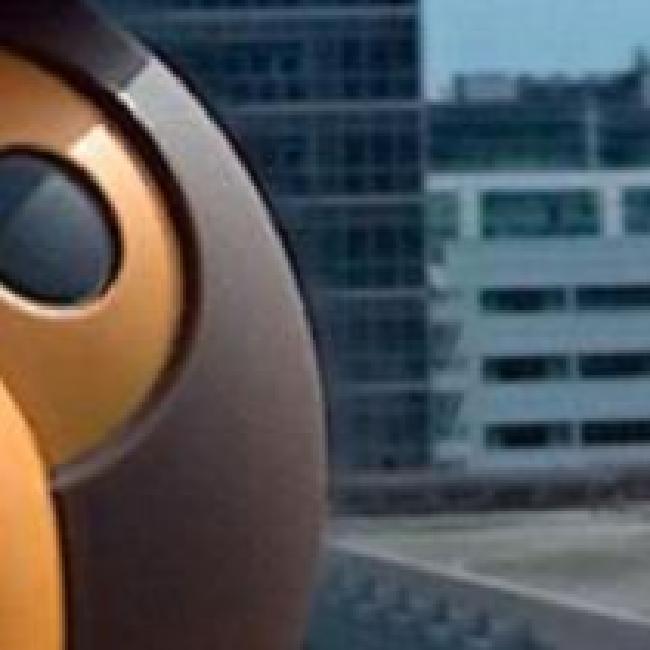}\\
Source A & Source B & All-in-focus (ours)\\
%(a)&(b)&(c)\\
\end{tabular}
\end{center}
\caption{Different focus source images and the all-in-focus resulting image. The sources A and B represent the same image in different focal planes. Our architecture combined these sources to create a new sharp image.}
\label{fig:initial}

\end{figure}

Most of the existing MFIF method contributions rely on proposals of new activity level measurements and/or fusion rules to solve the task. However, in recent years, this practice has been simplified through the employment of deep convolutional neural networks (CNN), and several deep learning-based methods have been introduced to create faster and simpler MFIF approaches.

In this paper, we address the MFIF problem also with a deep learning approach but with the novelty of using an end-to-end hourglass architecture to learn a direct mapping between source frames and the latent all-in-focus image. 
%To the best of our knowledge, this is the first time an hourglass network is applied to a multi-focus image fusion task. 
Our network intrinsically learns a focus map that contains the clarity information after comparing the pixel-wise sharpness of source images.

%The work most related to this research was proposed by Xiang Yan \textit{et al.} \cite{yan2018unsupervised}, which employs a structural similarity (SSIM) based loss function to achieve end-to-end unsupervised learning. Differently to our proposal, however, Xiang Yan \textit{et al.} use a Siamese-based architecture with several intermediate average fusions. This is a common approach in image fusion  \cite{liu2017multi,tang2018pixel} but it lacks flexibility when compared to multiple sources models where all frames are processed at the same time \cite{zagoruyko2015learning}. A drawback of \cite{yan2018unsupervised} is that test images from Lytro Multi-focus Image dataset \cite{nejati2015multi} were used for training, compromising the quality of reported results. Another method related to our approach is the segmentation-based model proposed by Liu \textit{et al.} \cite{liu2017multi}. 
%In their Siamese CNN method the multi-focus image fusion is treated as a pixel classification problem. However, the post-processing required to combine the classification of each patch from the image increases the total execution time (see Table \ref{tab:time}).
%In that case, the multi-focus image fusion is treated as a pixel classification problem and presented a fusion method based on a Siamese CNN. However, their work uses a patch by patch classification approach and subsequently a post-processing is required increasing the execution time.

To achieve the all-in-focus image, a Convolutional Neural Network with a encoder-decoder scheme trained with high quality images and their synthetically multi-focus blurred versions is adopted to obtain the mapping. This synthetic COCO multi-focus dataset is generated during training, providing almost infinite samples with no acquisition cost. The main novelty of this idea is the joint learning of the focus map and the fusion rule through a simple CNN model, which overcomes the typical complexity faced by existing fusion methods. Our method falls into the spatial domain category and is independent of frame size and amount of sources. The network constructed is also faster than most of the existing algorithms because regress the all-in-focus image at once, differently from the other traditional patch-based algorithms. Also, a commutative multiple sources model is adopted so fusion occurs equally independent of pair order.

%% file: related.tex
\section{Related Work}
% CHAMARIA ESSA SEÇAO DE BACKGROUND - GERMANO. RELATED WORK FICARIA ACIMA.

The work most related to this research was proposed by Xiang Yan \textit{et al.} \cite{yan2018unsupervised}, which employs a structural similarity (SSIM) based loss function to achieve end-to-end unsupervised learning. Differently to our proposal, however, Xiang Yan \textit{et al.} use a Siamese-based architecture with several intermediate average fusions. This is a common approach in image fusion  \cite{liu2017multi,tang2018pixel} but it lacks flexibility when compared to multiple sources models where all frames are processed at the same time \cite{zagoruyko2015learning}. A drawback of \cite{yan2018unsupervised} is that test images from Lytro Multi-focus Image dataset \cite{nejati2015multi} were used for training, compromising the quality of reported results. Another method related to our approach is the segmentation-based model proposed by Liu \textit{et al.} \cite{liu2017multi}. 
In their Siamese CNN method, the multi-focus image fusion is treated as a pixel classification problem. However, the post-processing required to combine the classification of each patch from the image increases the total execution time (see Table \ref{tab:time}).

\section{Background}
\label{sec:background}

Several methods have been proposed in the past for image fusion and, particularly, for multi-focus image fusion. Depending on the adopted fusion, the methods can be classified either as a transform domain or a spatial domain-based approach \cite{nejati2015multi}. While most methods fall into the first category, recent advances in neural networks have attracted the attention to spatial domain approaches, mostly due to performance improvements.

\textbf{Transform domain methods.} This class of method, such as in every transformation approach in computer vision, attempts to solve the problem in an alternate domain where finding the solution becomes simpler. In multi-focus methods, one usually transform the source images to a multi-scale domain, a subset of coefficients is selected or filtered from each source, and then a fusion of the decomposed coefficients is applied generating a reconstructed image in the corresponding domain. Finally, an inverse transform creates an all-in-focus spatial image. Main contributions in this area are in transformation selection, filtering of coefficients, and formulation of fusion rules. Some of the methods employ Gradient Pyramid \cite{petrovic2004gradient}, Wavelet Transforms \cite{lewis2007pixel}, Contourlet Transform \cite{zhang2009multifocus} and Discrete Cosine Transform \cite{haghighat2010real}, \cite{haghighat2011multi}. These methods usually have higher computational costs due to the transform and inverse transform operations. Some methods do not even specify the domain, but they try to learn the best feature space to solve the problem. Examples include the approaches based on Independent Component Analysis and Sparse Representation \cite{yang2010multifocus}.

\textbf{Spatial domain methods.} Differently to the previous approach, methods in this category try to reconstruct the all-in-focus image using intensity information. The formulation usually relies on the proposal of a focus metric that allows selecting the sharpest pixel within the sources. A sequence of filtering or morphological operations is also common in this kind of methods. Some of the most representative approaches include the Image Matting for fusion \cite{li2013image} and the Guided Filtering Fusion \cite{li2013image1}, both proposed by Li, Kang and Hu with results comparable to transform domain strategies but without the associated computational cost incurred by transformations. However, their manually designed morphological filtering assumes specific priors that may not apply to all images.

Recent spatial methods use deep learning as an alternative to handcrafted solutions \cite{liu2017multi,tang2018pixel,yan2018unsupervised}. Their main contributions are on the creation of network architecture and training datasets. Since the proposed architectures are generally Siamese based, these methods use a local neighborhood feature approach where every pixel is classified either as blurred or sharp. Despite the apparent good results, morphological post-processing is still needed to resolve global features, \eg\., filling holes. This increases the execution time as well as might add an unnecessary constraint to the solution space, no small holes, for example.
%this kind of strategy usually needs a morphological post-processing to include the global context information. 
%With this, the execution time is increased and they may also include constraint to the solution space. %The more recent approaches for MFIF are the CNN proposed by Liu \textit{et al.} \cite{liu2017multi}, the p-CNN proposed by Tang \textit{et al.} \cite{tang2018pixel} and the unsupervised method proposed by Yan \textit{et al.} \cite{yan2018unsupervised}.

%% file: method.tex
\section{Proposed Method}
\label{sec:method}

As mentioned above, we formulate the multi-focus image fusion problem as a multiple source segmentation/regression process where two frames are given to a Convolutional Heteroencoder and an RGB all-in-focus image is obtained.

\begin{figure}[t!]
\begin{center}
\includegraphics[width=.8\linewidth]{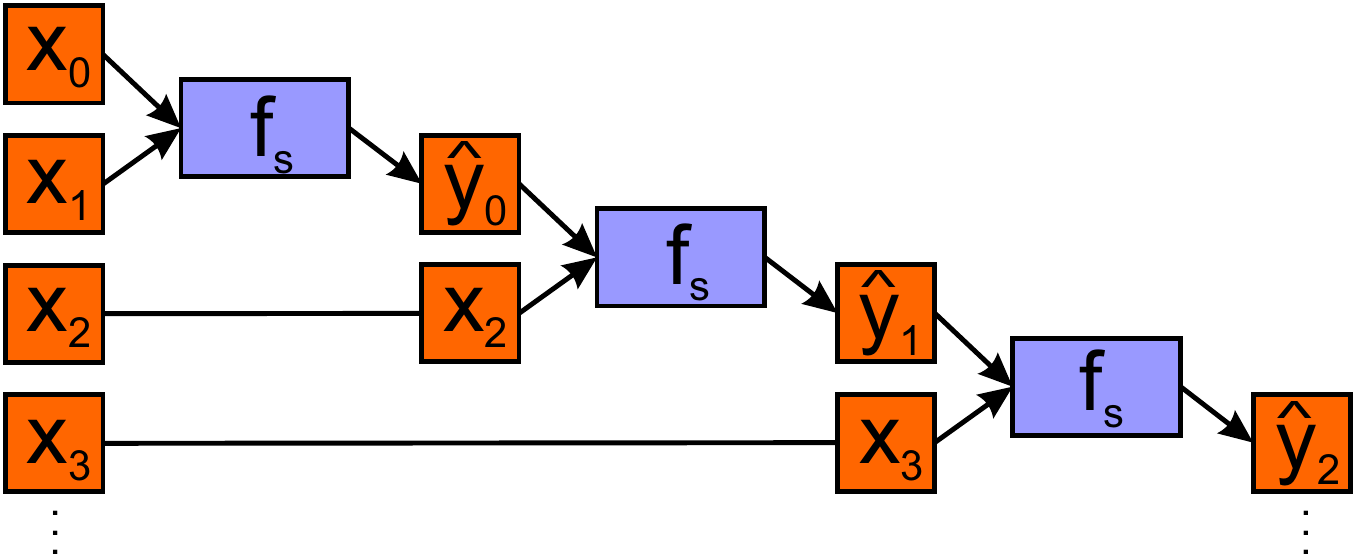}
    \end{center}
    \caption{Overall method scheme for a multi-focus fusion of an input burst. The images within the burst are incrementally fused through the $n$-th functional power $f^n$.}
    \label{fig:process}
\end{figure}

We defined the set of all multi-focus image pair as $\tuple{X}=\{ \tuple{x}_k\left|\right. \tuple{x}_k=(x_{kA},x_{kB})\}$, where $x_k\colon\Omega\to\Re{3}, \Omega\subset\Re{2}$, is an RGB source image. We are given a training set $S=\{(\tuple{x}_{0},y_0),\ldots, (\tuple{x}_{l},y_l)\}$, with cardinality $|S| = l$, where $\tuple{x}_k \in \tuple{X}$ is a source image pair and $y_k\colon\Omega\to\Re{3}$ is an all-in-focus ground truth image. 
%We call $g^0$ and $g^1$ the focused pixels subsets on source $x_A$ and $x_B$ respectively. We write the pixel indicator function $\indi{g^c}(p)$ simply as $v(p,c)$, i.e. $v(p,c) = 1$ if $p\in g^c$, otherwise $v(p,c) = 0$, with $c=\{0,1\}$. 
Let $\tuple{x}=(x_A,x_B)$ be a generic source tuple of $\tuple{X}$ and $y$ its focused ground truth. Our goal is to find a fusion function $f(\tuple{x})$ which takes two sources frames with different focus as input and obtain a fused image $\hat{y}$ as close as possible to the latent image $y$, $\hat{y}\approx y$. Note that a fusion function $f$ must be independent to pair order and therefore must meet the commutative law. This is regarded as $f(\tuple{x})=f(\bar{\tuple{x}})$ where $\bar{\tuple{x}}$ is the reverse order of the tuple $\tuple{x}$, $\bar{\tuple{x}}=(x_B,x_A)$. Function $f$ is then approximated here by U-Net\cite{ronneberger2015u}, a well-known hourglass architecture. We ensure the commutative property through an appropriate training protocol as described later. Although $f$ is bi-variable, a generalization for bursts $\tuple{x}^n=(x_0,...,x_n)$ with $n+1$ frames can be defined as the $n$-th functional power $f^n$, $f^n(\tuple{x}^{n})=(f \circ f^{n-1})(\tuple{x}^{n})$, where $\circ$ represent the partial composition operation, \eg, $f^2(\tuple{x}^2)=(f \circ f)(\tuple{x}^2)=f(f(x_0,x_1),x_2)$. Fig. \ref{fig:process} shows the overall process for multi-focus fusion of $n$ frames. A detailed explanation for every stage is given below.

%The fusion function is here defined through a hourglass segmentation network $f_h$, $f_h\colon\tuple{X}\to[0,1]^{|\Omega|}$, that learn to predict the focus map used in the fusion step. When $\tuple{x}$ is evaluated by $f_h$ a probability map $\tuple{z}\colon\Omega\to\Re{2}$ is obtained such that $\tuple{z}=(z_0,z_1)$ reports the probabilities of pixel $p \in \Omega$ be focus in source image $x_A$ and $x_B$ respectively. Without loss of generality let $z_0$ be the output probability map of $f_h(\tuple{x})$, and $\bar{z}_0$ be the output probability map of $f_h(\bar{\tuple{x}})$, been $\bar{\tuple{x}}$ the reverse order of the tuple $\tuple{x}$, $\bar{\tuple{x}}=(x_B,x_A)$. The unknown function $f_h$ meet that $z_0=1-z_1$, guaranteed through the logit function, but also must meet that $\bar{z}_0=z_1$ which indicates that evaluation of reversed order tuple $f_h(\bar{\tuple{x}})$ must be the exact complement of $f_h(\tuple{x})$, $f_h(\bar{\tuple{x}})=1-f_h(\tuple{x})$. This can be seen as the \textit{blurrest} trichotomous  relation ($\succ$) between corresponding pixels where only one of the following holds: pixel $p$ in $x_A$ is blurrer than in $x_B$ $\left(x_A(p) \succ x_B(p)\right)$, pixel $p$ in $x_B$ is blurrer than in $x_A$ ($x_B(p) \succ x_A(p)$) or pixel $p$ in $x_A$ is equal to $x_B$ ($x_A(p) = x_B(p)$). This property is ensured during $f_h$ training and $f$ fusion function computation.

\subsection{Dataset}
\label{sec:dataset}

As stated before, our target function $f$ is approximated through a CNN, and training such a neural network to predict the latent focused image given two blurry inputs requires a vast amount of training data. To the best of our knowledge there is no public multi-focus image fusion dataset with the all-in-focus ground truth available. Then, we synthetically generate our dataset to train the CNN. A potential idea would be to apply blur in some randomly selected patches of a sharp image $y$, and create the pair $\tuple{x}$ with the blurred and sharp patches, \eg, if $x_A$ is blurred then $x_B$ is its corresponding sharp patch from $y$. This approach was used recently by Liu \textit{et al.} \cite{liu2017multi} where the ISLR classification dataset was used to generate the training data. However, because our network is not a patch classification approach, the final input sources are required to contain a focus map where focused and blurred regions appear in the same frame. Following this idea the data generation method proposed in \cite{tang2018pixel} simulate situations where an image patch include both focused and de-focused regions. This is done defining 12 masks of blurred and unchanged areas used as focus map. Nevertheless, this small size set of masks might be insufficient to model the latent focus maps space significantly. Also, we find it very expensive to create an MFIF dataset by hand, given the enormous amount of ground truth data required to train the network.

Here, we propose to generate our dataset by applying synthetic blur to randomly selected objects instances extracted from the MS COCO segmentation dataset \cite{lin2014microsoft}. This dataset contains highly varied real-world images collected from the internet and its segmentation ground truth. Let $E=\{(y_0,g_0),\ldots,(y_m,g_m)\}$ be a panoptic segmentation set where $y_k$ is an image and $g_k$ is its segmentation mask, $g_k\colon\Omega\to\{0,\ldots,\gamma_k\}$ being $\gamma_k$ the amount of segmented objects. Let $(y,g)$ be a generic tuple from $E$ where there are $\gamma$ segmented objects. Let $\Gamma \subset \{0,\ldots,\gamma\}$ be a randomly selected subset of objects of $g$. Then, can be defined a focus map set $G=\{p\left|\right.c(p)\in\Gamma \}$ where $c(p)$ returns the object number assigned to pixel $p$, $c\colon\Omega\to\{0,\ldots,\gamma\}$. A binary focus map $g^b\colon\Omega\to\{0,1\}$, is then defined as $g^b(p)=\indi{G}(p)$ where $\indi{G}$ is the indicator function over $G$, \eg, $g^b(p)=1$ if $c(p)\in \Gamma$, otherwise $g^b(p)=0$.

A Gaussian blur kernel $h_\sigma$ is created using a uniform generated standard deviation $\sigma\sim U(1,5)$. Then, a blurred image $\bar{y}=y \ast h_\sigma$ is obtained by convolving the focused image with the blur kernel. Finally, a multi-focus input tuple $\tuple{x}=(x_A,x_B)$ is generated on-the-fly using the focus map $g^b$ and the blurred and sharp versions of the frame $y$ (Eq. \ref{eq:syntuple}).

\begin{equation}
\tuple{x}= \left(\bar{y} \cdot g^b+ y \cdot (1-g^b) ,\bar{y} \cdot (1- g^b)+ y \cdot g^b\right)
\label{eq:syntuple}
\end{equation}

\begin{figure}[tb!]
    \footnotesize
    \setlength{\tabcolsep}{1pt}
    \begin{tabular}{ccc}
    \includegraphics[width=.32\linewidth]{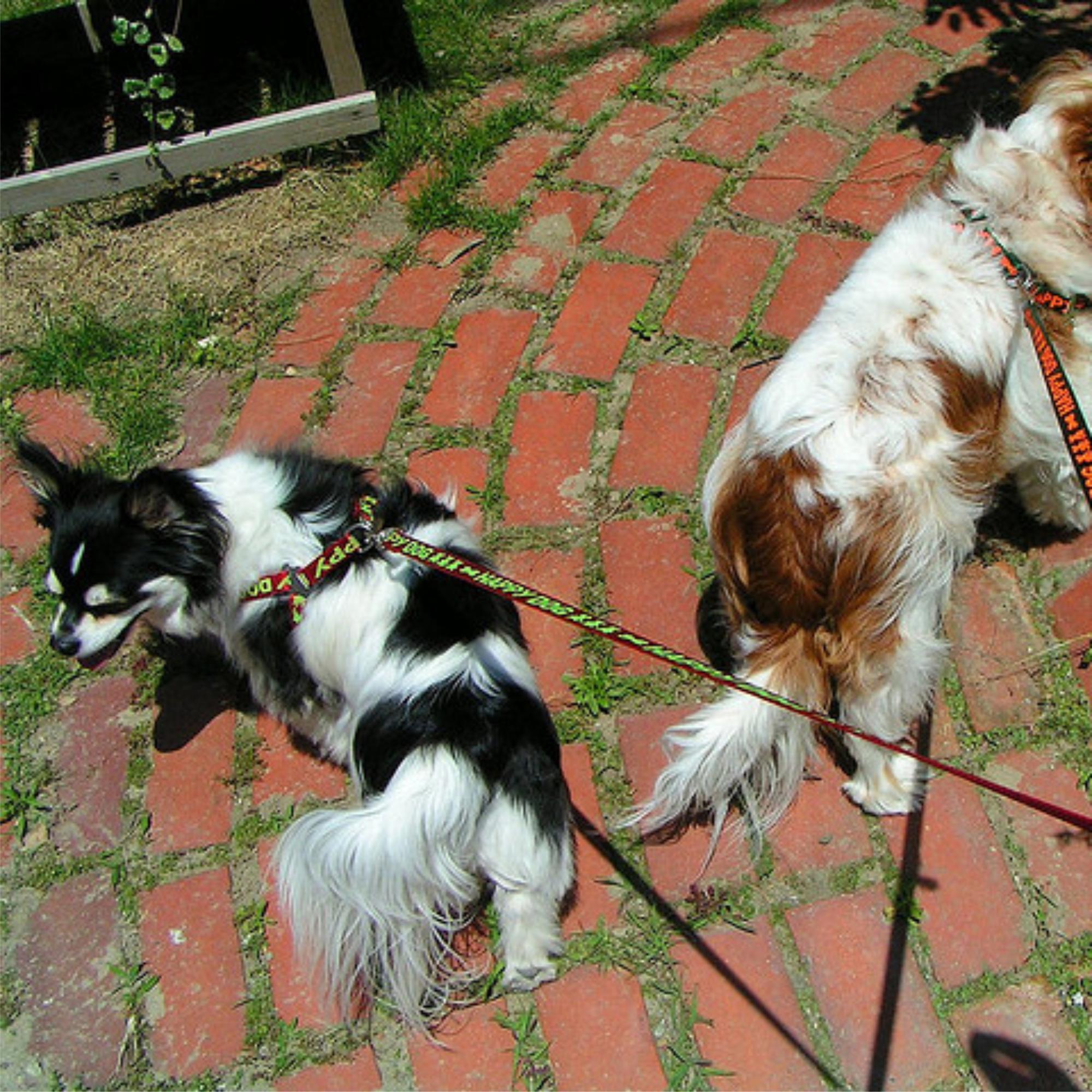}&
    \includegraphics[width=.32\linewidth]{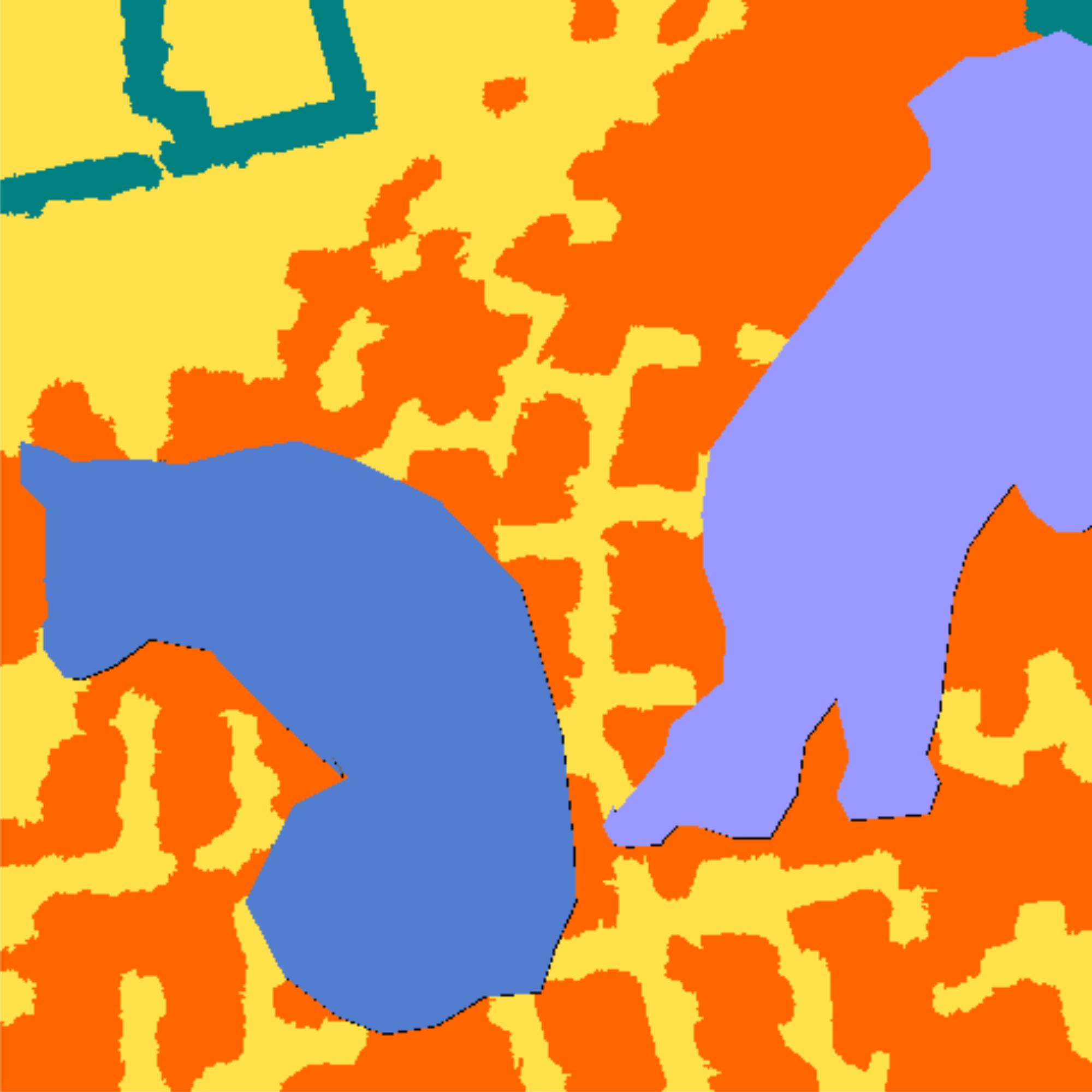}&
    \includegraphics[width=.32\linewidth]{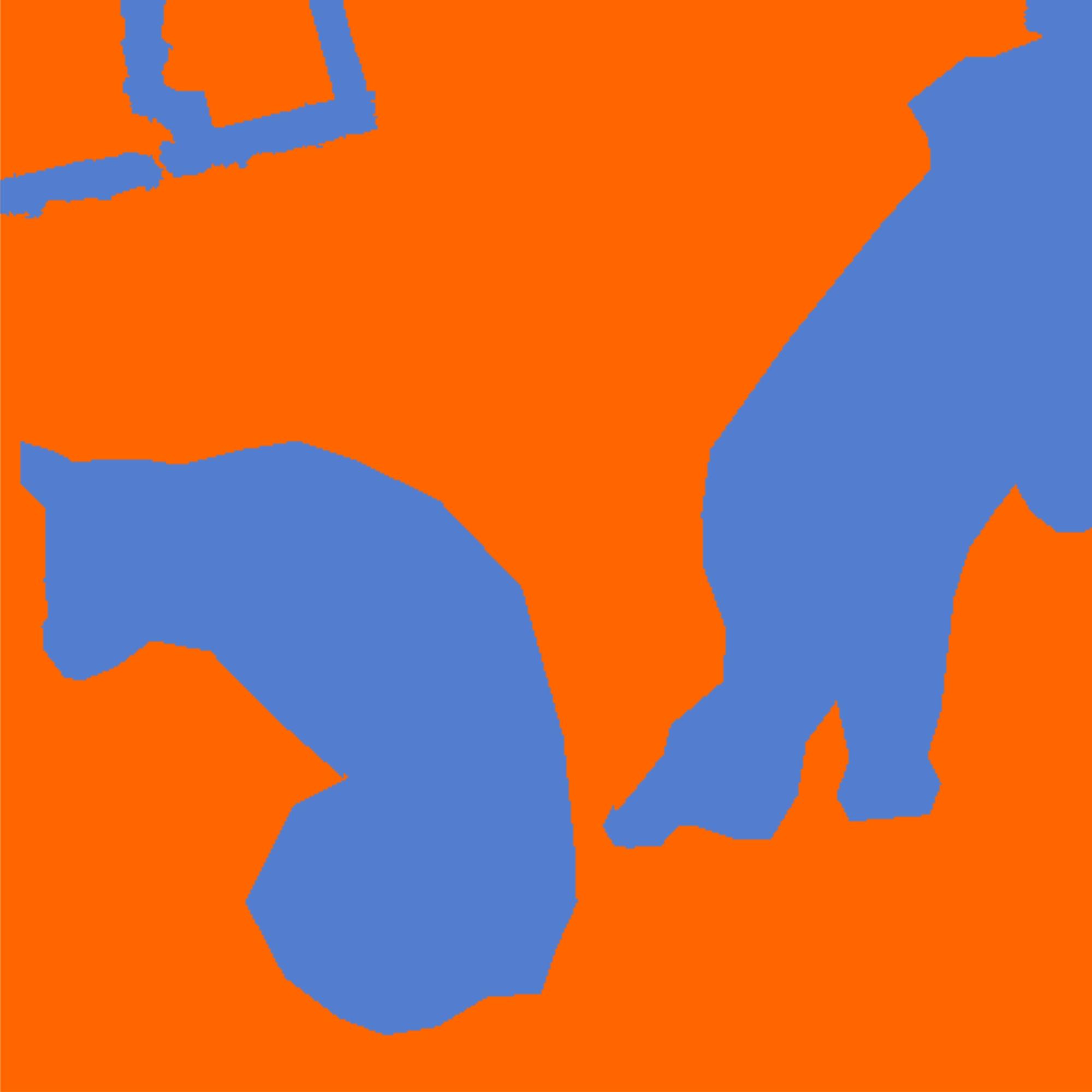}\\
    Image $y$&Segmentation mask $g$&Focus map $g^b$\\
    \includegraphics[width=.32\linewidth]{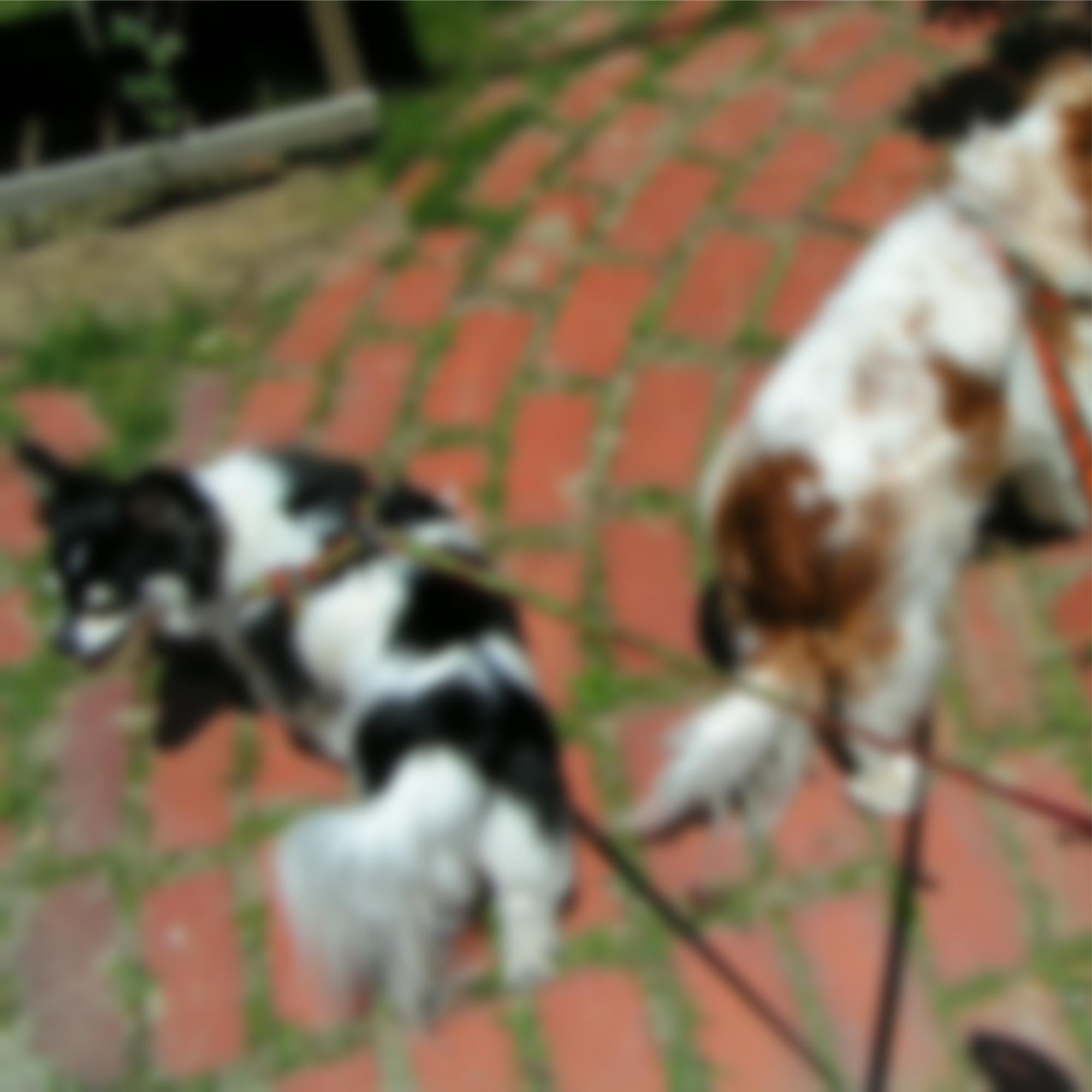}&
    \includegraphics[width=.32\linewidth]{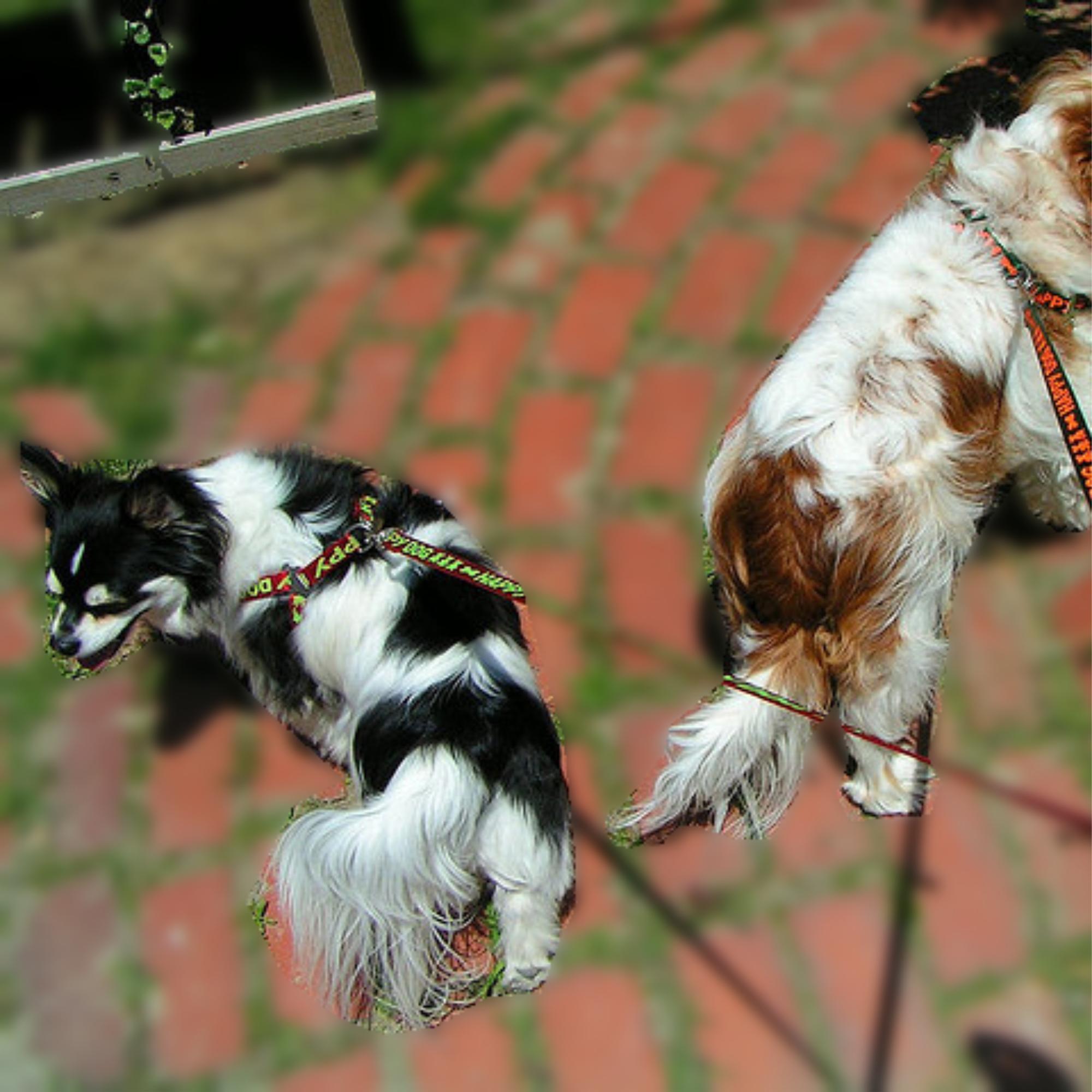}&
    \includegraphics[width=.32\linewidth]{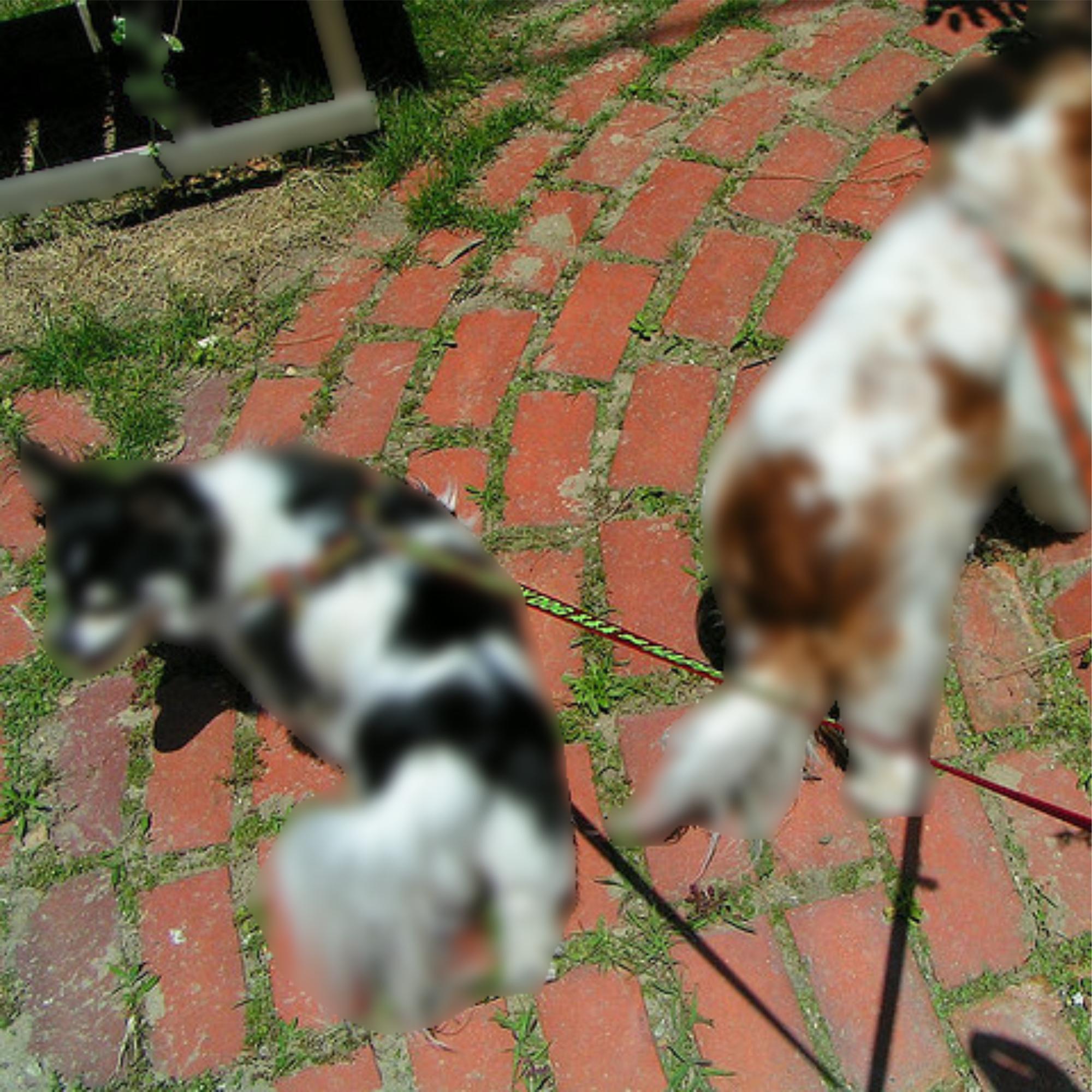}\\
    Blurred image $\bar{y}$&Source A  &Source B  \\
    &$\bar{y} \cdot g^b+ y \cdot (1-g^b)$&$\bar{y} \cdot (1- g^b)+ y \cdot g^b$\\
    \end{tabular}
    \caption{Example of synthetic tuple $\tuple{x}$ creation in our MFIF dataset using MS COCO image $y$ and its segmentation mask $g$. The focus map $g^b$ was created using two classes as background and the other three objects as foreground. The blurred image $\bar{y}$ and resulting sources $(x_A,x_B)$ are shown in the second row.}
    \label{fig:dataset}
\end{figure}

A generated sample of our realistic synthetic dataset is shown in Fig. \ref{fig:dataset}, with the corresponding sharp image $y$ and its segmentation mask $g$. Some objects randomly selected were taken as background leaving the rest in the foreground, resulting in the focus map $g^b$. Finally, the generated source frames are shown in the last row, computed according to Eq. \ref{eq:syntuple}. Hence, this approach gives nearly an infinite amount of training data. For fair evaluation, we employed the provided training and validation set split of the MS COCO dataset, and parameter optimization of the fusion network was applied to the training set only.

% \begin{figure*}[tb!]
%     \footnotesize
%     \setlength{\tabcolsep}{1pt}
%     \begin{tabular}{c}
%     \includegraphics[width=0.95\linewidth]{images/hgseg.eps}\\
%     a) HF Segmentation (HF-Seg) neural network. \\
%     \includegraphics[width=0.95\linewidth]{images/hgreg.eps}\\    
%     b) HF Regression (HF-Reg) neural network. \\
%     \end{tabular}
%     \caption{Multiple sources hourglass networks for multi-focus image fusion using a) segmentation and weighted average fusion and b) a regression method with a learned fusion rule. Sources are showed separated in the figures but the input block is 6 channels depth.}
%     \label{fig:hgreg}
% \end{figure*}

\begin{figure*}[tb!]
    \footnotesize
    \setlength{\tabcolsep}{1pt}
    \begin{tabular}{c}
    \includegraphics[width=0.95\linewidth]{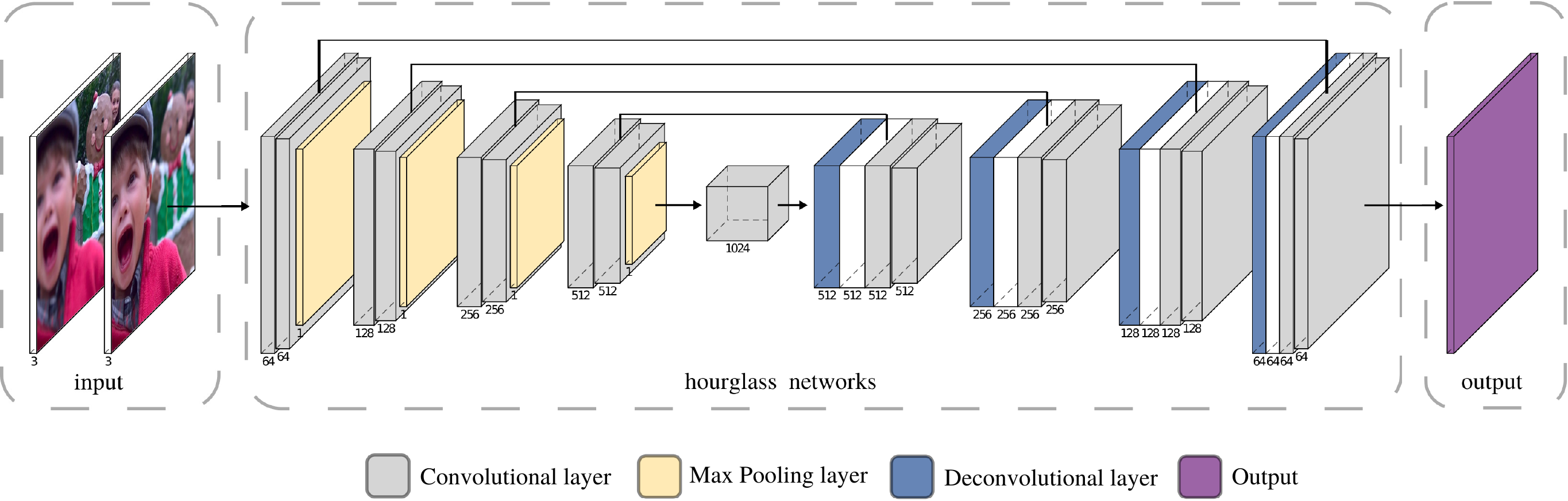}\\
    \end{tabular}
    \caption{Multiple sources hourglass networks for multi-focus image fusion. Sources are showed separated in the figures but the input block is 6 channels depth. For HF-Reg the output layer corresponds to the sharp image estimate. In the case of HF-Seg, the output layer is a 2-channel feature map, and each channel $z_i$ represents the probability of selecting the input source $i$. }
    \label{fig:net}
\end{figure*}

\subsection{Multiple Sources Hourglass Network}
%[explain unet, approach with segmentation and manual fusion, approach with regression and automatic fusion, loss functions,commutative property]

To approximate the fusion function $f$ we explored two ideas in the U-Net hourglass architecture. This is an encoder-decoder type of network where the first half of the layers contracts the width and heights of feature maps increasing the analyzed field of view. A significantly smaller representation is learned in the deepest block, forcing the identification of sufficiently relevant features to describe the inputs. Then, the second half of the layers acts as a reconstruction path leading to a feature space with the same width and height of those of the source inputs. Skipping connections linking the same depths in the encoder and decoder branches are used to localize and propagate high resolution features. The network does not have any fully connected layers and only uses the valid part of each convolution, \eg, the output map only contains the pixels, for which the full context is available in the input image \cite{ronneberger2015u}. An extension for multiple input sources is proposed here based on the results of \cite{zagoruyko2015learning} to learn a similarity function. The superiority of multiple source approaches was validated when compared with Siamese methods, which takes a single image as input in the feature extraction path. Nevertheless, we generalize here the scheme proposed by Zagoruyko \textit{et. al} for single value regression, to full RGB images regression/segmentation tasks. 

We present two variants of hourglass architectures to solve the multi-focus image fusion problem: HF-Seg and  HF-Reg. \figurename~\ref{fig:net} show general architecture.

% \subsubsection{Fusion map prediction}

% \textbf{Fusion map prediction.}  Our first proposal uses the hourglass network for fusion map estimation. This is based on the ideas of \cite{liu2017multi, tang2018pixel} for obtaining a focus map. Differently to theirs, here the problem is cast as a segmentation process where our HF-Seg architecture (Fig. \ref{fig:hgreg} a) receive two RGB sources as a 6-channels map $\tuple{x}=(x_A,x_b)$, and outputs, after the Softmax layer, a two channel segmentation map $\tuple{z}=(z_0,z_1)$. In practice, this segmentation map represents the predicted fusion map and its complement, $z_0=1-z_1$. After obtaining the focus map, the resulting fused image can be inferred by using a fusion rule. To this end a Fusion layer is added at the top of the network to obtain the all-in-focus image given the two input frames and the predicted focus map. The fusion function $f_S$ is here expressed as the pixel-wise weighted-average rule of the network output \cite{liu2017multi,tang2018pixel}:

\textbf{Fusion map prediction (HF-Seg).} Our first proposal uses the hourglass network for fusion map estimation. This is based on the ideas of \cite{tang2018pixel, liu2017multi} for obtaining a focus map. Differently to theirs, here the problem is cast as a segmentation process where our HF-Seg architecture (Fig. \ref{fig:net} receive two RGB sources as a 6-channels map $\tuple{x}=(x_A,x_b)$, and outputs is a Softmax layer, using for obtain two-channel segmentation map $\tuple{z}=(z_0,z_1)$. In practice, this segmentation map represents the predicted fusion map and its complement, $z_0=1-z_1$. After obtaining the focus map, the resulting fused image can be inferred by using a fusion rule. The fusion function $f_S$ is here expressed as the pixel-wise weighted-average rule of the network output \cite{liu2017multi,tang2018pixel}:

% To this end a Fusion layer is added at the top of the network to obtain the all-in-focus image given the two input frames and the predicted focus map.

\begin{equation}
f_S(\tuple{x})= z_0\cdot x_A + z_1\cdot x_B
\label{eq:frule}
\end{equation}

Training of such a network requires the ground truth of the fusion map for every input pair $\tuple{x}$ to be known. However, during the generation of the synthetic source, the focus map $g^b$ is obtained. Then, the HF-Seg training is carried out by only using the Binary Cross Entropy (BCE) loss function:

\begin{multline}
\mathcal{L}_S(\tuple{z},g^b)=- \frac{1}{|\Omega|}\sum_{p \in \Omega} g^b(p)\cdot \log\left(z_0(p)\right)\\ + \left(1-g^b(p)\right)\cdot \log\left(z_1(p)\right) 
\label{eq:lossbce}
\end{multline}

where $\tuple{z}=(z_0,z_1)$ is the output of HF-Seg and $g^b$ is created as described in Section \ref{sec:dataset}. 
% Note that for network training the Fusion layer is not needed. 

\begin{table*}[t!]
\begin{equation}
\mathcal{L}_R(\hat{y},y)=\frac{1}{|\Omega|}\sum_{i=0}^2 \sum_{p \in \Omega} \varphi_{\alpha}(y_i(p),\hat{y}_i(p)) + \sum_{i=0}^2 |\min(y_i) -\min(\hat{y}_i)| + \sum_{i=0}^2 |\max(y_i)-\max(\hat{y}_i)|
\label{eq:lossnsl1}
    \end{equation}
\end{table*}

% \subsubsection{All-in-focus image regression}

\textbf{All-in-focus image regression (HF-Reg).}  Although the HF-Seg approach is straightforward, the fusion rule has to be previously established (Eq. \ref{eq:frule}). Then, this network works better in problems where a focus map and a fusion rule can be used, such as in multi-focus image fusion. However, a more general model can be derived from the HF-Seg method to learn the best fusion rule for source combination automatically. This second proposal uses an end-to-end approach where the hourglass network is used to regress the all-in-focus image directly. Here, the fusion function input is also a 6-channels map. The architecture remains as a sequence of convolutions and max-pooling in the encoder and convolutions-upsampling blocks in the decoder. Differently to the previous approach, the output feature block is a 3-channel map $\hat{y}$ corresponding to an RGB focused image. 
% And, in this case, there are no Softmax and Fusion layers at the end of the HF-Reg (Fig. \ref{fig:hgreg}). Because the final image is directly regressed, the fusion function is jointly learned through the HF-Reg, $f_R(\tuple{x})=\hat{y}$. 
In this approach, the learning process requires an appropriate regression loss function rather than the BCE. Let in this context $y=(y_0,y_1,y_2)$ be a ground truth image where $y_0$, $y_1$ and $y_2$ are its RGB channels respectively. Similarly, the estimated RGB all-in-focus image is given by $\hat{y}=(\hat{y}_0,\hat{y}_1,\hat{y}_2)$. Our regression loss function is defined as in Eq. \ref{eq:lossnsl1}, where $\varphi_{\alpha}$ is an intensity dissimilarity function. Note that our loss is the sum of the mean distance for each channel, rather than the mean distance of all channels. This per-channel loss has shown to be better for color estimation because averaging the errors of the three channels usually lead to grayscale output space. Also, when values are regressed, the output space during training is not bounded as opposed to the previous segmentation approach. This lack of boundaries can bring difficulties to get an output map in the expected range. To this end, a regularization term that forces the convergence of minimum and maximum values of each channel was added to the loss function. This regularization term penalizes more severely fused images with low contrast or intensity values outside the interval $[0,1]$, assuring the output map to be in the right range in earlier training steps.

Among all dissimilarity functions available in the literature such as the Mean Square Error (MSE) and L1 norm, here we define $\varphi_{\alpha}$ as the Normalized Positive Sigmoid (NPS) between two intensities parameterized by $\alpha$:

\begin{equation}
\varphi_{\alpha}(y,\hat{y})= \frac{2}{e^{-\alpha\cdot|y-\hat{y}|}+1}-1=\frac{e^{\alpha\cdot|y-\hat{y}|}-1}{e^{\alpha\cdot|y-\hat{y}|}+1}
\label{eq:iexpl1}
\end{equation}

\begin{figure}[t!]
\small
\def\svgwidth{0.95\linewidth}
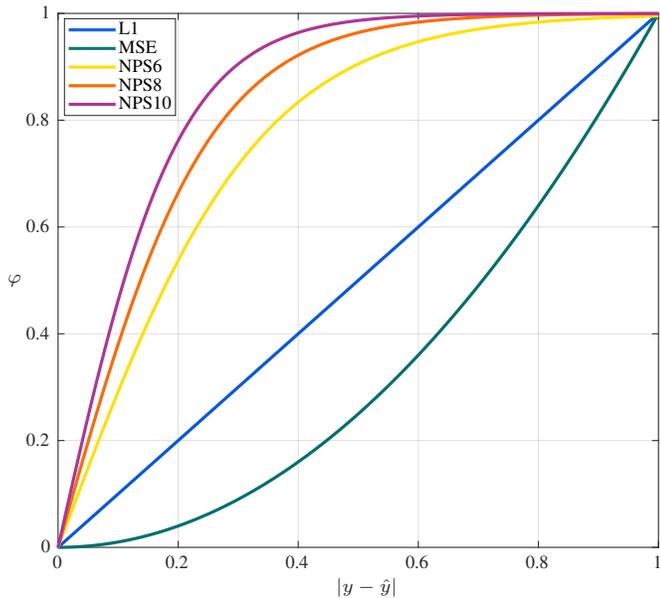
\caption{Distances mapping for L1, MSE, NPS6, NPS8 and NPS10 dissimilarity functions.}
    \label{fig:dist}
\end{figure}

Given the ground truth intensity $y$ and the estimated intensity $\hat{y}$, the minimum metric value is obtained when $\hat{y}=y$, $\varphi_{\alpha}(y,y)=0$. Also, the maximum value is approximately 1 for $\alpha > 5$, $\displaystyle \lim_{|y-\hat{y}| \to \infty} \varphi_{\alpha}= 1$. However, with our NPS, a lower decay is observed when compared to the usual MSE and L1 approaches. This behavior forces the propagation of higher errors, even with small intensity differences. Fig. \ref{fig:dist} shows the error mapping for the L1 norm, MSE, and NPS for different values of $\alpha$, \eg, NPS6, NPS8 and NPS10 corresponding to $\alpha=$ 6, 8 and 10, respectively. 

The simplicity of our network allows us to perform image fusion without further post-processing. Also, our approach is faster than most of state-of-the-art methods for MFIF. The network learns the best fusion function, and it is not limited to problems where the fusion map can be obtained, \eg, multi-modal fusion, and multi-exposure fusion.

\subsection{Implementation details}

To fulfill the commutative law, required for all fusion functions, an appropriate training protocol was employed. For every generated tuple $\tuple{x}=(x_A,x_B)$, we also forward in the same minibatch the inversed tuple $\bar{\tuple{x}}=(x_B,x_A)$. In the HF-Reg network training, any further ground truth modification for $\bar{\tuple{x}}$ is needed, because the all-in-focus image $y$ remains the same. However, for the HF-Seg approach, the ground truth focus map needs to be inverted, \eg, $1-g^b$, so the obtained reconstruction remains as close as possible to $y$.

Because the best pixel value that can be obtained belongs to one of the sources, \eg, the multi-focus image fusion problem can be seen as a selection problem where $y(p)$ is either equal to $x_A(p)$ or $x_B(p)$, a posterior post-processing for selecting the nearest value can be applied. Let $\hat{y}$ be a fused image obtained by $f_R(\tuple{x})$. The final all-in-focus image is obtained as follows:

\begin{equation}
\hat{y}^*(p)= \begin{cases}
x_A(p) & \textbf{ if } ||\hat{y}(p) - x_A(p)||^2 <  ||\hat{y}(p) - x_B(p)||^2 \\
x_B(p) & \text{ otherwise } \\
\end{cases}
\end{equation}

%% file: images/nps.pdf_tex
%% Accompanies image file 'nps.pdf' (pdf, eps, ps)
\begingroup%
  \makeatletter%
  \providecommand\color[2][]{%
    \errmessage{(Inkscape) Color is used for the text in Inkscape, but the package 'color.sty' is not loaded}%
    \renewcommand\color[2][]{}%
  }%
  \providecommand\transparent[1]{%
    \errmessage{(Inkscape) Transparency is used (non-zero) for the text in Inkscape, but the package 'transparent.sty' is not loaded}%
    \renewcommand\transparent[1]{}%
  }%
  \providecommand\rotatebox[2]{#2}%
  \newcommand*\fsize{\dimexpr\f@size pt\relax}%
  \newcommand*\lineheight[1]{\fontsize{\fsize}{#1\fsize}\selectfont}%
  \ifx\svgwidth\undefined%
    \setlength{\unitlength}{403.83701925bp}%
    \ifx\svgscale\undefined%
      \relax%
    \else%
      \setlength{\unitlength}{\unitlength * \real{\svgscale}}%
    \fi%
  \else%
    \setlength{\unitlength}{\svgwidth}%
  \fi%
  \global\let\svgwidth\undefined%
  \global\let\svgscale\undefined%
  \makeatother%
  \begin{picture}(,0.88732389)%
    \lineheight{1}%
    \setlength\tabcolsep{0pt}%
    \put(0,0){\includegraphics[width=\unitlength,page=1]{images/nps.pdf}}%
    \put(0.0407978,0.00037143){\color[rgb]{0.14901961,0.14901961,0.14901961}\makebox(0,0)[lt]{\lineheight{1.25}\smash{\scriptsize\begin{tabular}[t]{l}0\end{tabular}}}}%
    \put(0.21875115,0.00037143){\color[rgb]{0.14901961,0.14901961,0.14901961}\makebox(0,0)[lt]{\lineheight{1.25}\smash{\scriptsize\begin{tabular}[t]{l}0.2\end{tabular}}}}%
    \put(0.40847836,0.00037143){\color[rgb]{0.14901961,0.14901961,0.14901961}\makebox(0,0)[lt]{\lineheight{1.25}\smash{\scriptsize\begin{tabular}[t]{l}0.4\end{tabular}}}}%
    \put(0.59820555,0.00037143){\color[rgb]{0.14901961,0.14901961,0.14901961}\makebox(0,0)[lt]{\lineheight{1.25}\smash{\scriptsize\begin{tabular}[t]{l}0.6\end{tabular}}}}%
    \put(0.78793272,0.00037143){\color[rgb]{0.14901961,0.14901961,0.14901961}\makebox(0,0)[lt]{\lineheight{1.25}\smash{\scriptsize\begin{tabular}[t]{l}0.8\end{tabular}}}}%
    \put(0.98943375,0.00037143){\color[rgb]{0.14901961,0.14901961,0.14901961}\makebox(0,0)[lt]{\lineheight{1.25}\smash{\scriptsize\begin{tabular}[t]{l}1\end{tabular}}}}%
    \put(0.48820555,-0.037143){\color[rgb]{0.14901961,0.14901961,0.14901961}\makebox(0,0)[lt]{\lineheight{1.25}\smash{\scriptsize\begin{tabular}[t]{l}$|y-\hat{y}|$\end{tabular}}}}%
    \put(0,0){\includegraphics[width=\unitlength,page=2]{images/nps.pdf}}%
    \put(0.02248293,0.02541429){\color[rgb]{0.14901961,0.14901961,0.14901961}\makebox(0,0)[lt]{\lineheight{1.25}\smash{\scriptsize\begin{tabular}[t]{l}0\end{tabular}}}}%
    \put(-0.00106477,0.19428493){\color[rgb]{0.14901961,0.14901961,0.14901961}\makebox(0,0)[lt]{\lineheight{1.25}\smash{\scriptsize\begin{tabular}[t]{l}0.2\end{tabular}}}}%
    \put(-0.00106477,0.36315557){\color[rgb]{0.14901961,0.14901961,0.14901961}\makebox(0,0)[lt]{\lineheight{1.25}\smash{\scriptsize\begin{tabular}[t]{l}0.4\end{tabular}}}}%
    \put(-0.00106477,0.53202629){\color[rgb]{0.14901961,0.14901961,0.14901961}\makebox(0,0)[lt]{\lineheight{1.25}\smash{\scriptsize\begin{tabular}[t]{l}0.6\end{tabular}}}}%
    \put(-0.00106477,0.70089692){\color[rgb]{0.14901961,0.14901961,0.14901961}\makebox(0,0)[lt]{\lineheight{1.25}\smash{\scriptsize\begin{tabular}[t]{l}0.8\end{tabular}}}}%
    \put(0.02248293,0.86976757){\color[rgb]{0.14901961,0.14901961,0.14901961}\makebox(0,0)[lt]{\lineheight{1.25}\smash{\scriptsize\begin{tabular}[t]{l}1\end{tabular}}}}%
    \put(-0.0306477,0.45202629){\color[rgb]{0.14901961,0.14901961,0.14901961}\makebox(0,0)[lt]{\lineheight{1.25}\smash{\scriptsize\begin{tabular}[t]{l}$\varphi$\end{tabular}}}}%
    \put(0,0){\includegraphics[width=\unitlength,page=3]{images/nps.pdf}}%
    \put(0.14521618,0.84453789){\color[rgb]{0,0,0}\makebox(0,0)[lt]{\lineheight{1.25}\smash{\scriptsize\begin{tabular}[t]{l}L1\end{tabular}}}}%
    \put(0,0){\includegraphics[width=\unitlength,page=4]{images/nps.pdf}}%
    \put(0.14521618,0.81510328){\color[rgb]{0,0,0}\makebox(0,0)[lt]{\lineheight{1.25}\smash{\scriptsize\begin{tabular}[t]{l}MSE\end{tabular}}}}%
    \put(0,0){\includegraphics[width=\unitlength,page=5]{images/nps.pdf}}%
    \put(0.14521618,0.78566863){\color[rgb]{0,0,0}\makebox(0,0)[lt]{\lineheight{1.25}\smash{\scriptsize\begin{tabular}[t]{l}NPS6\end{tabular}}}}%
    \put(0,0){\includegraphics[width=\unitlength,page=6]{images/nps.pdf}}%
    \put(0.14521618,0.75623402){\color[rgb]{0,0,0}\makebox(0,0)[lt]{\lineheight{1.25}\smash{\scriptsize\begin{tabular}[t]{l}NPS8\end{tabular}}}}%
    \put(0,0){\includegraphics[width=\unitlength,page=7]{images/nps.pdf}}%
    \put(0.14521618,0.72679937){\color[rgb]{0,0,0}\makebox(0,0)[lt]{\lineheight{1.25}\smash{\scriptsize\begin{tabular}[t]{l}NPS10\end{tabular}}}}%
    \put(0,0){\includegraphics[width=\unitlength,page=8]{images/nps.pdf}}%
  \end{picture}%
\endgroup%

%% file: results.tex
\section{Experiments}

To evaluate and validate our hourglass fusion networks were conducted several experiments. For comparison were used the Image Matting for fusion (IM) \cite{li2013image}, the variance-based image fusion in DCT domain (DCT) \cite{haghighat2010real} and with consistency verification (DCT+CV) \cite{haghighat2011multi}, the Guided Filtering Fusion (GFF) \cite{li2013image1} and the deep Convolutional Neural Network (CNN) \cite{liu2017multi} approaches. We refer as Near the nearest source color post-processing explained in the previous section. The experiments were conducted over synthetic and real datasets with different amount of images within the burst.

We trained both networks over a synthetic multi-focus dataset as stated before. Was used the optimizer Adam \cite{kingma2014adam} with its defaults parameters and the initial learning rate was set to $10^{-5}$. The number of epochs and minibatch size was $1000$ and $3$ respectively. For training purpose were applied random crops of $400 \times 400$ and mirroring. Networks initialization was made with normally distributed weights using Xavier's method \cite{glorot2010understanding}. For the test phase was used the size of the original images since after learning the kernels the networks are size invariant.

\subsection{Commutativity}

\begin{figure}[t!]
    \footnotesize
    \setlength{\tabcolsep}{1pt}
    \begin{tabular}{cc}
      
    %\multicolumn{2}{c}{Tuple $\tuple{x}_1$}&
    \multicolumn{2}{c}{Tuple $\tuple{x}_1$}\\
    \includegraphics[width=.49\linewidth]{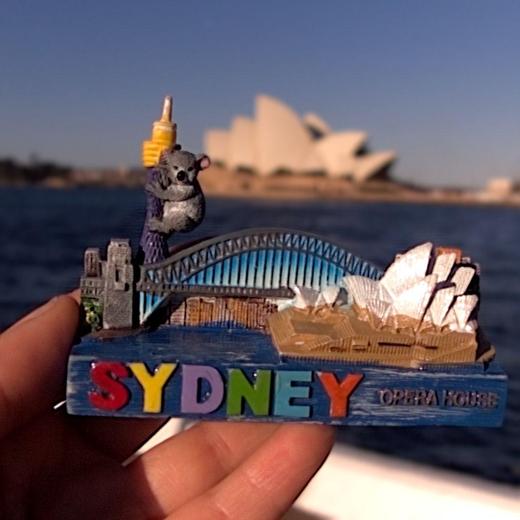}&  
    \includegraphics[width=.49\linewidth]{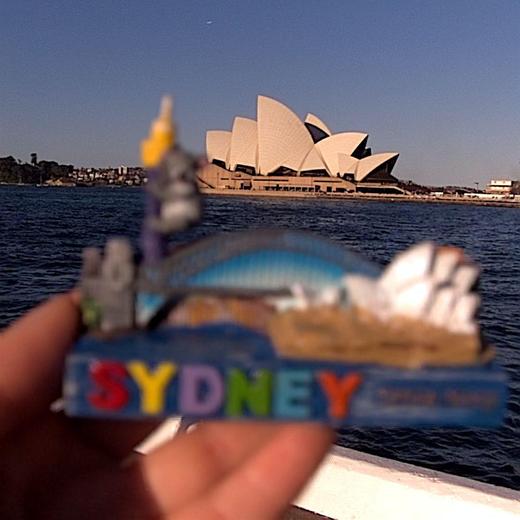}\\ 
    
    %Output $f_R(\tuple{x}_1)$&Output $f_R(\bar{\tuple{x}}_1)$&
    Output $f_R(\tuple{x}_1)$&Output $f_R(\bar{\tuple{x}}_1)$\\
    \includegraphics[width=.49\linewidth]{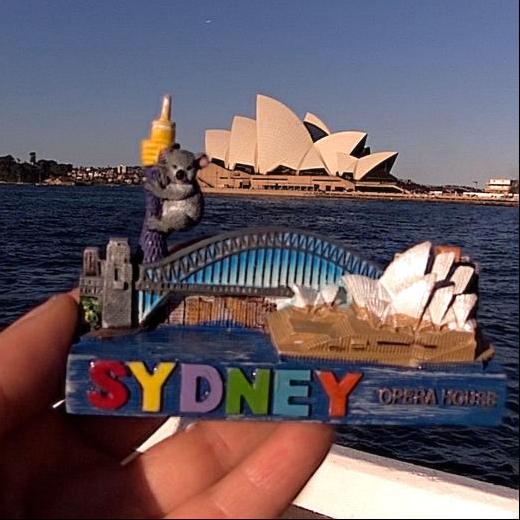}&
    \includegraphics[width=.49\linewidth]{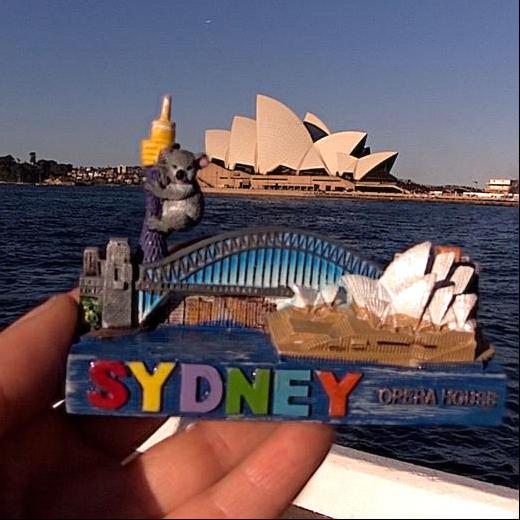}\\
    
    %Output $f_S(\tuple{x}_1)$&Output $f_S(\bar{\tuple{x}}_1)$&
    Output $f_S(\tuple{x}_1)$&Output $f_S(\bar{\tuple{x}}_1)$\\
    \includegraphics[width=.49\linewidth]{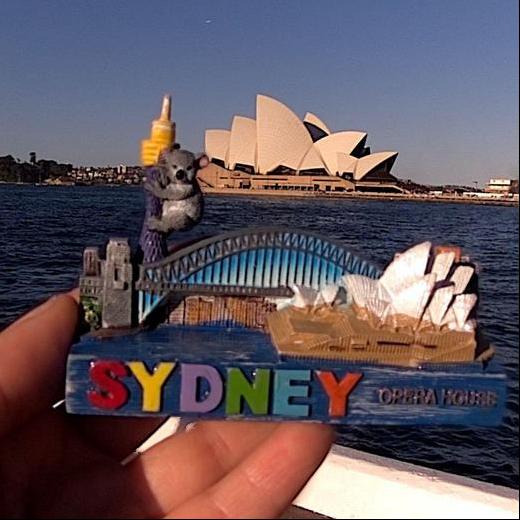}&
    \includegraphics[width=.49\linewidth]{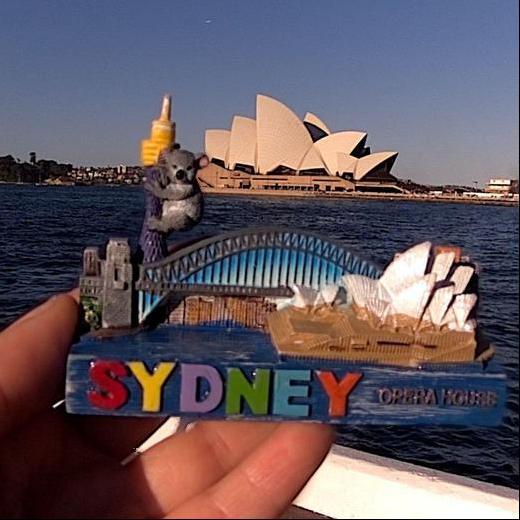}\\  
    
    %$|f_R(\tuple{x}_1)-f_R(\bar{\tuple{x}}_1)|$&$|f_S(\tuple{x}_1)-f_S(\bar{\tuple{x}}_1)|$&$|f_R(\tuple{x}_2)-f_R(\bar{\tuple{x}}_2)|$&$|f_S(\tuple{x}_2)-f_S(\bar{\tuple{x}}_2)|$\\
    %\includegraphics[width=.24\linewidth]{images/3fusion_regdif.jpg}&
    %\includegraphics[width=.24\linewidth]{images/3fusion_segdif.jpg}&  
    %\includegraphics[width=.24\linewidth]{images/13fusion_regdif.jpg}&
    %\includegraphics[width=.24\linewidth]{images/13fusion_segdif.jpg}\\
    
    \end{tabular}
    
    \caption{Example of fusion results for a tuple with normal and reversed order. In the first row are shown the frames within the tuples. In second and third row are shown the fusion results with the regression and segmentation networks respectively for both normal and reverse order evaluations.}
    \label{fig:rescommutativity}
\end{figure}

As stated before, all fusion functions must meet that no matter the order of the sources, the all-in-focus image must remain. Because the hourglass network input is a six-channel map,  $\tuple{x}$ and $\bar{\tuple{x}}$ are different objects, and therefore, the output might be different. However, due to the training protocol detailed in previous sections, the learned fusion function leads to approximately the same point in the output space for inputs $\tuple{x}$ and $\bar{\tuple{x}}$, ensuring the required commutative property. Fig. \ref{fig:rescommutativity} shows two different pairs $\tuple{x}$ from the real dataset, and the results obtained doing a forward of the tuple and its reverse into each proposed network. 
%Difference map between every output of the same network are also shown in the figure. 
As can be seen no significant differences are observed in the all-in-focus images. 
%The error image color map was encoded such that blue means error 0 and red colors are error 1. 
The obtained mean squared error between $f(\tuple{x})$ and $f(\bar{\tuple{x}})$ was in the order of $10^{-5}$ for all images and can not be visually perceived. The property remained for all tested images.

\subsection{MFIF metrics}

\begin{figure*}[tb!]
    \footnotesize
    \setlength{\tabcolsep}{1pt}
    \begin{tabular}{|c|c|c|c|c|}
    \hline
    &HF-Reg (without Near) & HF-Reg (with Near) &Dummy A& Dummy B\\
    &
    \includegraphics[width=.23\linewidth]{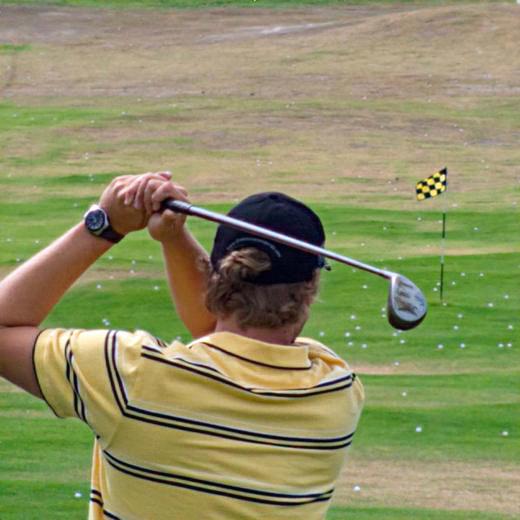}&
    \includegraphics[width=.23\linewidth]{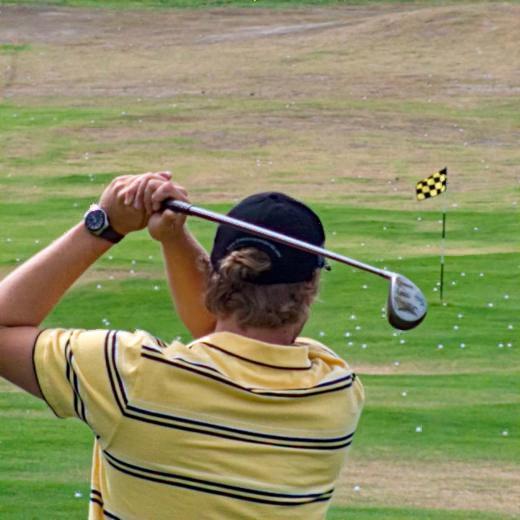}&
    \includegraphics[width=.23\linewidth]{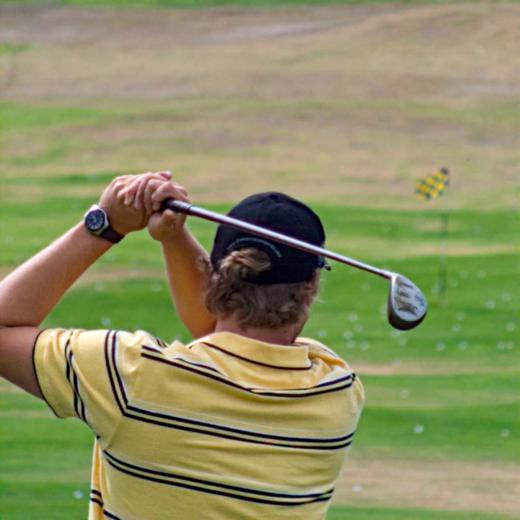}&    
    \includegraphics[width=.23\linewidth]{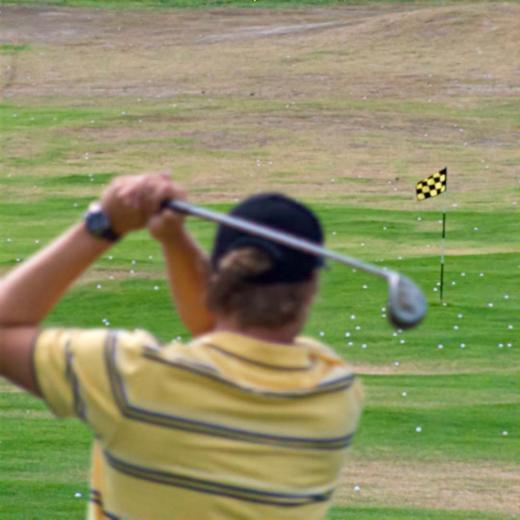}\\ \hline
    
    $Q_{MI}$  & 0.8463 & 1.1097 &\textbf{1.2812} & \textbf{1.2812} \\ \hline
    $Q_{TE}$  & 0.3616 & 0.3766 &\textbf{0.4432} & \textbf{0.4435} \\ \hline
    $Q_{NCIE}$& 0.8212 & 0.8336 &\textbf{0.8631} & \textbf{0.8628} \\ \hline
    $Q_G$  & 0.6255 & \textbf{0.6768} &0.5330 & 0.6614 \\ \hline 
    $Q_P$   & 0.7151 & 0.7610 & 0.7210 & \textbf{0.8007} \\ \hline
    $Q_S$   & \textbf{0.9510} & 0.9473 &0.8536 & 0.8841 \\ \hline
    $Q_{CB}$  & 0.7336 & \textbf{0.7806} &0.6955 & 0.7591 \\ \hline
    \end{tabular}
    
    \caption{Example of the values of the fusion metrics for HF-Reg without and with Near post-processing and two dummy methods that returns the first (Dummy A) and second (Dummy B) image of the tuple as result for the fusion.}
    \label{fig:badmetrics}
\end{figure*}

Quantitative evaluation analysis for image fusion problems is a challenging task since the reference all-in-focus images are not known. Among the several proposals introduced in the literature, it is challenging to select which one is the best. We explore some of the most used metrics like Normalize Mutual Information $Q_{MI}$, Tsallis Entropy $Q_{TE}$, Nonlinear Correlation Information Entropy $Q_{NCIE}$, Gradient-based $Q_G$, Phase Congruency $Q_P$, Piella-Heijmans $Q_S$, and Chen-Blum $Q_{CB}$. We follow $Q_{MI}$ Hossny definition because it reduces the bias of the original $Q_{MI}$ metric toward the sources. Every metric belongs to one of the four groups of objective assessment metrics, information theory, feature-based, structural similarity-based, and human perception inspired. Higher metrics values mean better fusion quality. A detailed explanation of each metric can be found in \cite{liu2012objective}. Despite the generalized use of this metrics, we found that computing the agreement of the resulting image with every source, including blurred regions of the sources, may not represent a proper measurement of the fusion quality.  Liu \textit{et al.} \cite{liu2012objective} also arrive at this conclusion in their work "The lack of IQM-to-MIF metric correlation is because most fusion metrics count on how the input images are fused together rather than the quality of the fused image. Note: When the input images are of significantly different quality, we found that a fusion metric may lead to a confused judgment."

An example of bias toward the source is shown in Fig. \ref{fig:badmetrics}. The first image in the figure refers to the output of our HF-Reg network without Near post-processing, followed by the same image after the nearest post-processing. Dummy A and Dummy B images correspond with the outputs of the methods that return exactly the source A and B, respectively.  As can be seeing in the figure, most metrics get higher values when the output are one of the sources. This means that a dummy method that outputs an input image will get a better metric value than others that returns a visually acceptable all-focused image. The behavior is expected because most of the metrics find a quality value using the similarity between the resulting image with each source. Then, when an all-in-focus image is obtained with subtle colors variation respect to sources, the metrics values highly decrease as in the case of our HF-Reg network without Near. The values for dummies methods even super-passes most of the literature methods, so caution must be taken when using objective assessment metrics to give a conclusive result. We also compute the full reference Structural SIMilarity index (SSIM) between the resulting fused image and the all-in-focus ground truth in the synthetic dataset.

\subsection{L1 vs MSE vs NPS}

\begin{figure}[tb!]
    \footnotesize
    \setlength{\tabcolsep}{1pt}
    \begin{tabular}{ccc}
    \multicolumn{3}{c}{L1}\\
    \includegraphics[width=.32\linewidth]{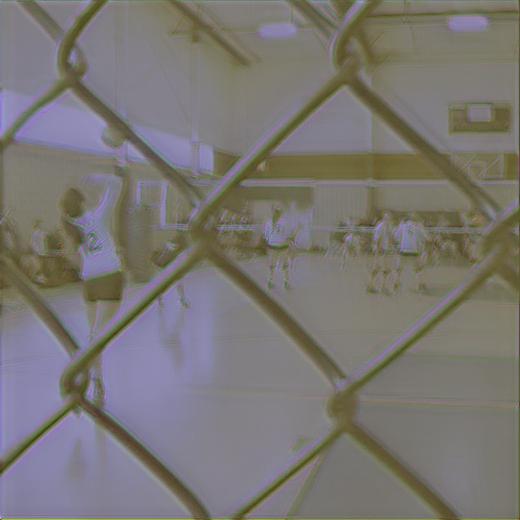}&    
    \includegraphics[width=.32\linewidth]{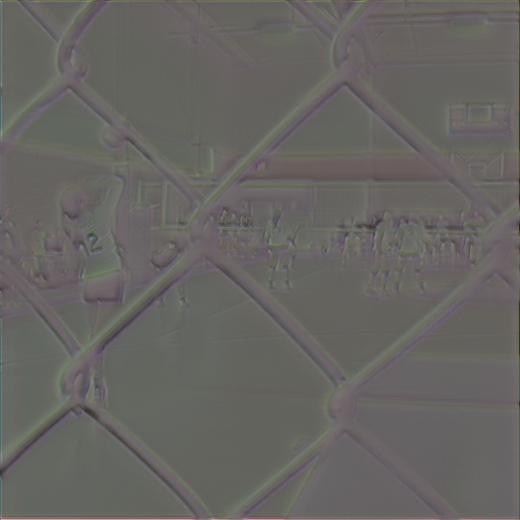}&
    \includegraphics[width=.32\linewidth]{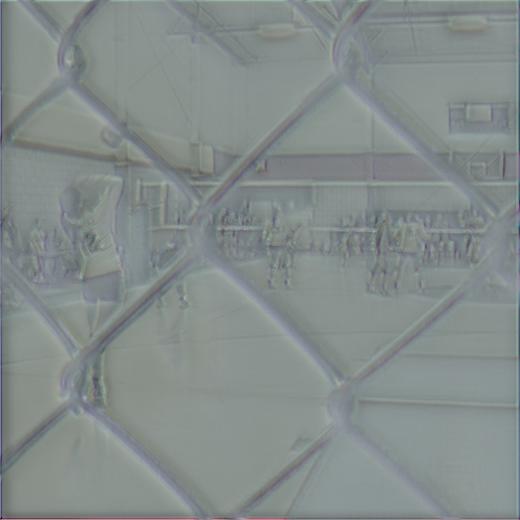}\\
    
    \multicolumn{3}{c}{MSE}\\
    \includegraphics[width=.32\linewidth]{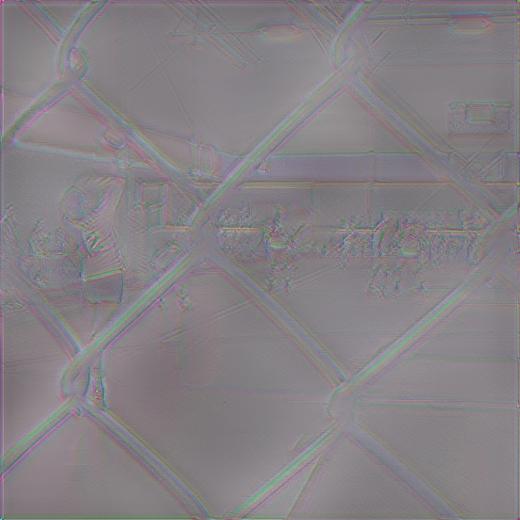}&    
    \includegraphics[width=.32\linewidth]{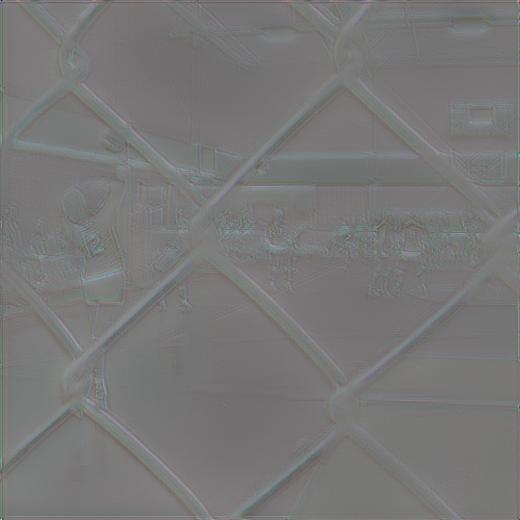}&
    \includegraphics[width=.32\linewidth]{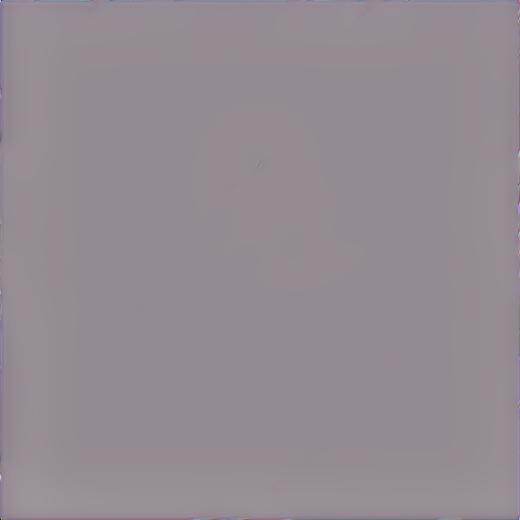}\\
    
    \multicolumn{3}{c}{NPS6}\\
    \includegraphics[width=.32\linewidth]{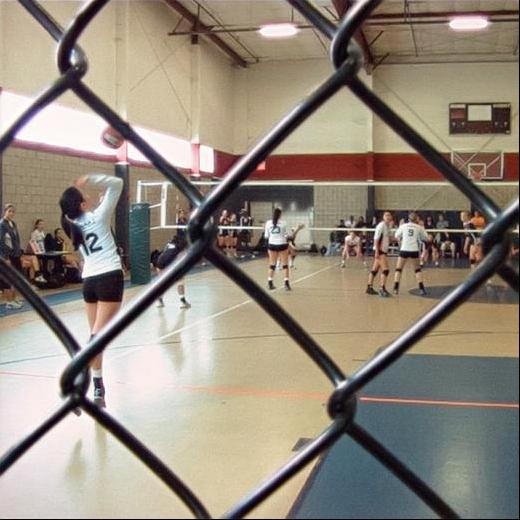}&    
    \includegraphics[width=.32\linewidth]{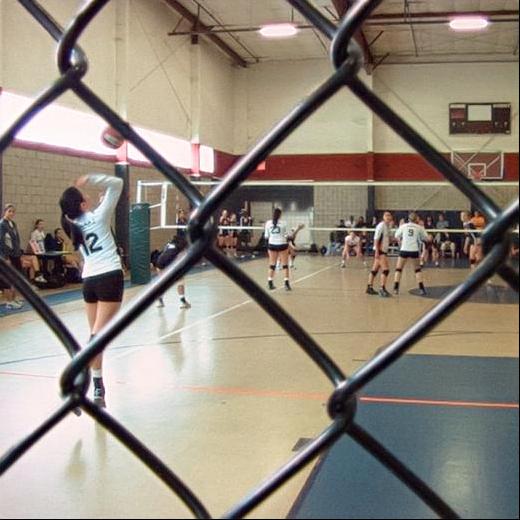}&
    \includegraphics[width=.32\linewidth]{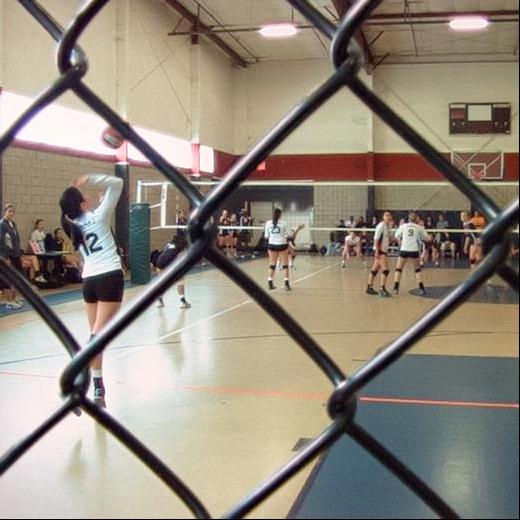}\\
    
    Epoch 200 & Epoch 600  & Epoch 1000\\
    %Epoch 200 & Epoch 400 & Epoch 600 & Epoch 800 & Epoch 1000\\
    
    \end{tabular}
    
    \caption{Example of the multi-focus image fusion obtained with intermediate L1, MSE and HF-Reg networks during the training.}
    \label{fig:npsvsall}
\end{figure}

The first experiment has the objective to show the feasibility of our NPS loss function. The HF-Reg architecture was trained over the synthetic COCO multi-focus dataset but using L1, MSE and NPS6. Every 20 epochs the weights of the network were saved. Training hyper-parameters are the same described at the beginning of the section. After training during 1000 epochs, a synthetic multi-focus test dataset was created for evaluation purpose. This dataset is composed of 100 randomly selected images from the test data of the COCO panoptic segmentation, and then the multi-focus data creation previously described was applied. Despite the usefulness of L1 and MSE loss functions in other regression problems, we founded difficult to regress the appropriated all-in-focus image. The obtained output during different epochs of the training is shown in Fig. \ref{fig:npsvsall} for every training function over a real image from the Lytro dataset. No consistent learning was observed when used L1 or MSE loss function. However, with our NPS6 loss function the colors and contrast of the regressed image are well estimated even in earlier epochs. 

\begin{figure}[t!]
\small
\def\svgwidth{0.95\linewidth}
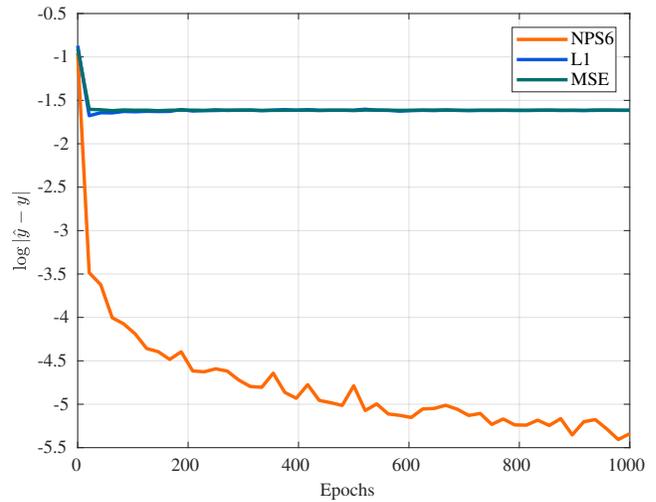
\caption{Logarithm of the error over the synthetic multi-focus test dataset.}
    \label{fig:loss}
\end{figure}

The behavior is corroborated by the mean errors curve over the synthetic dataset (Fig. \ref{fig:loss}). This figure shows the mean L1 difference between estimated all-in-focus image $\hat{y}$ and the ground truth $y$ over different epochs. The y-axis is shown in $\log$ scale for better interpretation. With our NPS we succeed to obtain a visually good solution for the MFIF problem, and the error curve trending suggests that if further training is performed an even lower error can be obtained.

\subsection{Two Source synthetic dataset}

\begin{figure}[t!]
\small
\def\svgwidth{0.95\linewidth}
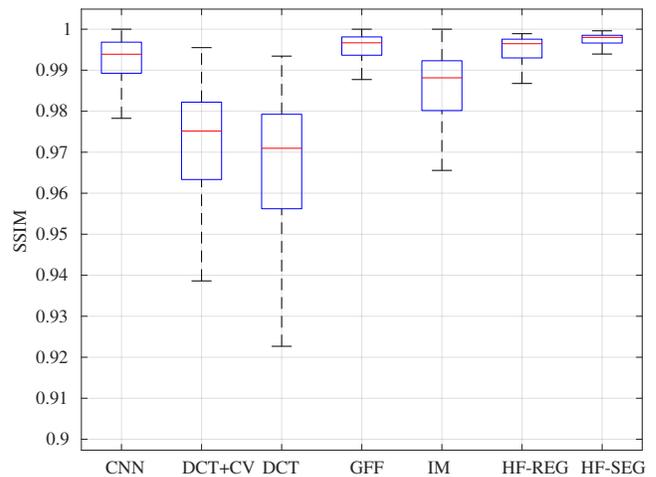
\caption{Box plot for SSIM reference metric over the synthetic multi-focus test dataset.}
    \label{fig:syn_ssim}
\end{figure}

For evaluation purpose, we validate our method in the synthetic multi-focus test dataset. The dataset has 100 pairs with its corresponding all-in-focus ground truth. Because the reference image is known, the SSIM metric between the obtained reconstruction and the ground truth was used in the evaluation. Fig. \ref{fig:syn_ssim} shows the obtained box plots over the SSIM metric for every tested method. As can be observed, a high mean with a small variance is seen in our HF-Seg approach that has nearly $1$ SSIM for most of the pairs. Our HF-Reg also behaves well obtaining comparable results to GFF and lower variance respect to CNN. In three of the seven objective assessment metrics, our methods have higher mean and lower variance than the methods in the state-of-the-art (Table \ref{tab:syn_metrics}). However, despite the higher mean value in some references and multi-focus metrics, no statistically significant difference was measured for the results of the CNN, GFF, HF-Reg, and HF-Seg according to the Friedman test and Nemenyi post-hoc.

\begin{table*}
\scriptsize
\caption{Mean and standard deviation of the objective assessment over the synthetic multi-focus test dataset.}
\label{tab:syn_metrics}
\resizebox{\linewidth}{!}{%
\begin{tabular}{cccccccccc}
Metrics& CNN \cite{liu2017multi}& DCT+CV \cite{haghighat2011multi}&DCT \cite{haghighat2010real}&GFF \cite{li2013image1}&IM \cite{li2013image}&HF-Reg (Ours)&HF-Seg (Ours)\\\hline
$Q_{MI}$&$ 1.1467 \pm 0.1474 $ & $ 0.9014 \pm 0.1777 $ & $ 0.8827 \pm 0.1726 $ & $ 1.0920 \pm 0.1695 $ & $ 1.1350 \pm 0.1501 $ & $ 1.1828 \pm 0.1107 $ & $ \mathbf{1.1924 \pm 0.1156} $ \\ 
$Q_{TE}$&$ 0.4101 \pm 0.0412 $ & $ 0.3869 \pm 0.0458 $ & $ 0.3810 \pm 0.0452 $ & $ 0.4049 \pm 0.0417 $ & $ 0.4055 \pm 0.0418 $ & $ 0.4129 \pm 0.0328 $ & $ \mathbf{0.4152 \pm 0.0352} $ \\ 
$Q_{NCIE}$&$ 0.8425 \pm 0.0111 $ & $ 0.8275 \pm 0.0100 $ & $ 0.8263 \pm 0.0092 $ & $ 0.8390 \pm 0.0121 $ & $ 0.8418 \pm 0.0113 $& $ 0.8432 \pm 0.0087 $ & $ \mathbf{0.8445 \pm 0.0095} $ \\ 
$Q_G$&$ 0.7499 \pm 0.0405 $ & $ 0.6788 \pm 0.0620 $ & $ 0.6759 \pm 0.0615 $ & $ \mathbf{0.7526 \pm 0.0390} $ & $ 0.7365 \pm 0.0447 $ & $ 0.6714 \pm 0.0895 $ & $ 0.7182 \pm 0.0548 $ \\ 
$Q_P$&$ \mathbf{0.7985 \pm 0.0816} $ & $ 0.7376 \pm 0.0870 $ & $ 0.6959 \pm 0.0963 $ & $ 0.7964 \pm 0.0811 $ & $ 0.7426 \pm 0.0831 $ & $ 0.7414 \pm 0.1110 $ & $ 0.7722 \pm 0.0938 $ \\ 
$Q_S$&$ 0.9566 \pm 0.0159 $ & $ 0.9411 \pm 0.0211 $ & $ 0.9408 \pm 0.0210 $ & $ \mathbf{0.9586 \pm 0.0144} $ & $ 0.9440 \pm 0.0210 $ & $ 0.9493 \pm 0.0174 $ & $ 0.9548 \pm 0.0153 $ \\ 
$Q_{CB}$&$ \mathbf{0.8198 \pm 0.0383} $ & $ 0.7112 \pm 0.0621 $ & $ 0.6838 \pm 0.0666 $ & $ 0.8125 \pm 0.0376 $ & $ 0.7950 \pm 0.0515 $ & $ 0.7449 \pm 0.0844 $ & $ 0.7719 \pm 0.0572 $ \\ 
\end{tabular}
}
\end{table*}

% \begin{figure}[tb!]
%     \footnotesize
%     \setlength{\tabcolsep}{1pt}
%     \begin{tabular}{ccc}
%     Source A& Source B& CNN\\
%     \includegraphics[width=.33\linewidth]{images/image099-A.jpg}&    
%     \includegraphics[width=.33\linewidth]{images/image099-B.jpg}&    
%     \includegraphics[width=.33\linewidth]{images/fused_cnn-100.jpg}\\ 
%     DCT+CV& DCT& GFF\\
%     \includegraphics[width=.33\linewidth]{images/fused_dct1-100.jpg}&
%     \includegraphics[width=.33\linewidth]{images/fused_dct2-100.jpg}&
%     \includegraphics[width=.33\linewidth]{images/fused_gff-100.jpg}\\
%     IFM& HF-Reg& HF-Seg\\
%     \includegraphics[width=.33\linewidth]{images/fused_ifm-100.jpg}&
%     \includegraphics[width=.33\linewidth]{images/2fusion_reg99.jpg}&
%     \includegraphics[width=.33\linewidth]{images/2fusion_seg99.jpg}\\
    
%     \end{tabular}
    
%     \caption{Results.}
%     \label{fig:synres1}
% \end{figure}

\begin{figure}[tb!]
    \footnotesize
    \setlength{\tabcolsep}{1pt}
    \begin{tabular}{ccc}
    Source A& Source B& CNN\\
    \includegraphics[width=.32\linewidth]{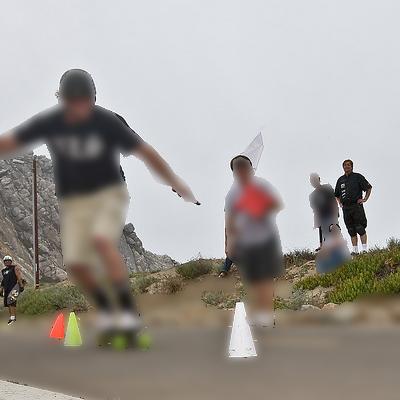}&    
    \includegraphics[width=.32\linewidth]{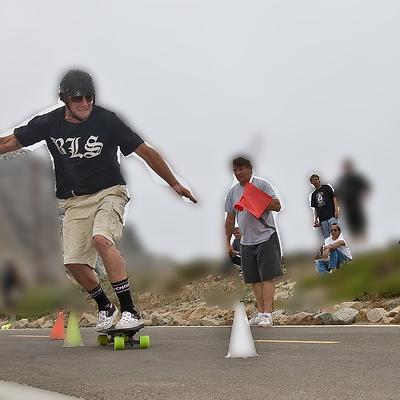}&    
    \includegraphics[width=.32\linewidth]{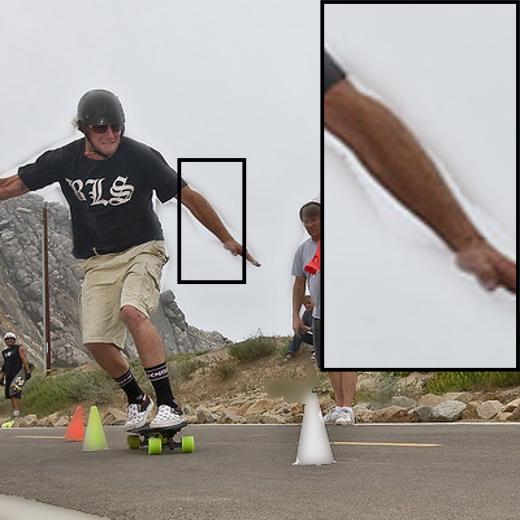}\\ 
    DCT+CV& DCT& GFF\\
    \includegraphics[width=.32\linewidth]{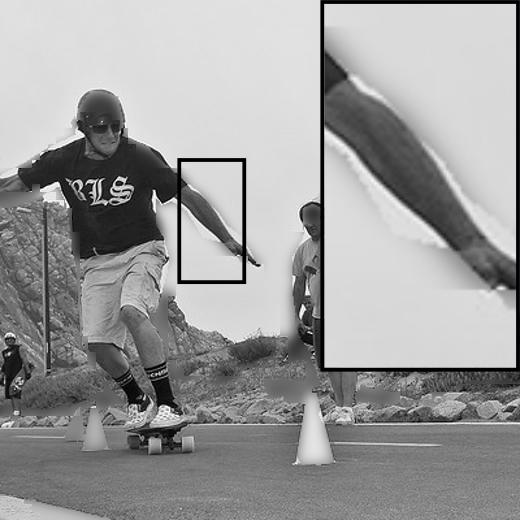}&
    \includegraphics[width=.32\linewidth]{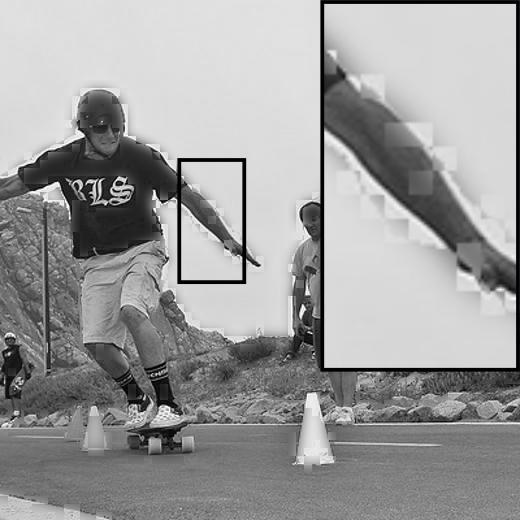}&
    \includegraphics[width=.32\linewidth]{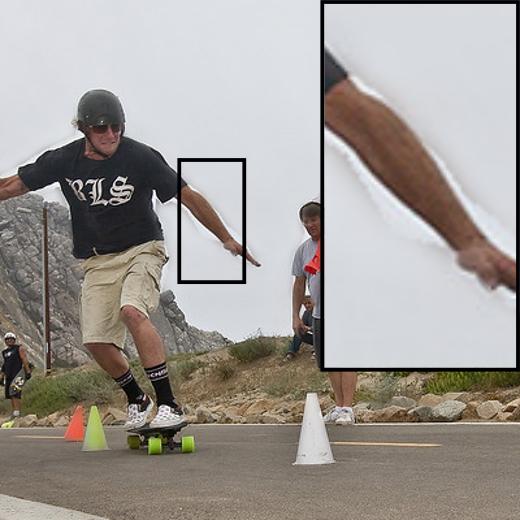}\\
    IM& HF-Reg& HF-Seg\\
    \includegraphics[width=.32\linewidth]{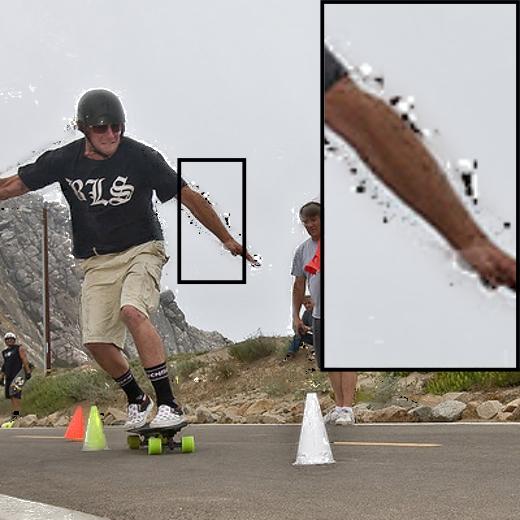}&
    \includegraphics[width=.32\linewidth]{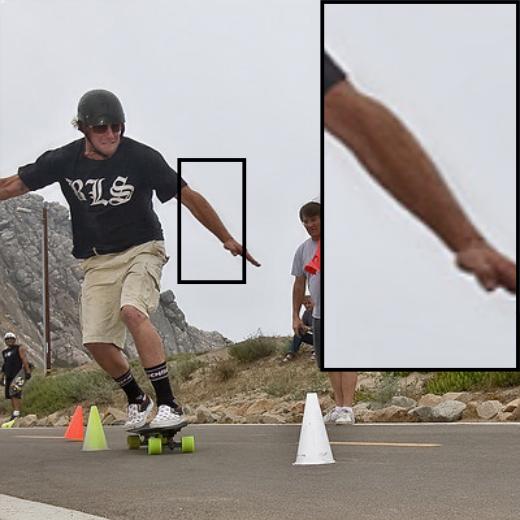}&
    \includegraphics[width=.32\linewidth]{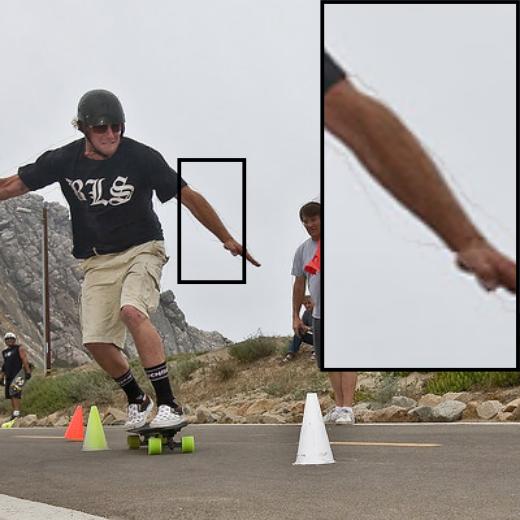}\\
    
    \end{tabular}
    
    \caption{Synthetic test example where most methods have higher values in objective assessment metrics than ours, but a better quality fusion is obtained with the proposals.}
    \label{fig:synres2}
\end{figure}

\begin{figure}[t!]
\small
\def\svgwidth{0.95\linewidth}
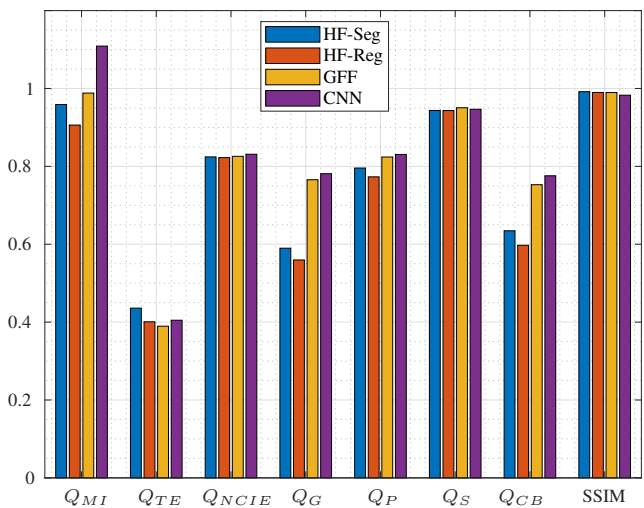
\caption{Values of the quality metrics for images in Fig. \ref{fig:synres2}.}
    \label{fig:syn_metrics}
\end{figure}

For almost every pair in the synthetic test dataset the CNN, GFF, HF-Reg and HF-Seg approaches returns a similar focused image with very few differences in term of pixels colors. However, as stated before, sometimes the metrics can confuse the judgment of the fusion quality as in the example shown in Fig. \ref{fig:synres2}. For this pair CNN and GFF outperform our approaches for most metrics except $Q_{TE}$ and SSIM (Fig. \ref{fig:syn_metrics}) but, as can be seen, our networks outputs a better quality fusion result. 

\subsection{Two Sources real dataset}

\begin{figure}[b!]
    \footnotesize
    \setlength{\tabcolsep}{1pt}
    \begin{tabular}{cccc}
    Source A& Source B&HF-Reg&HF-Seg\\

    \includegraphics[width=.24\linewidth]{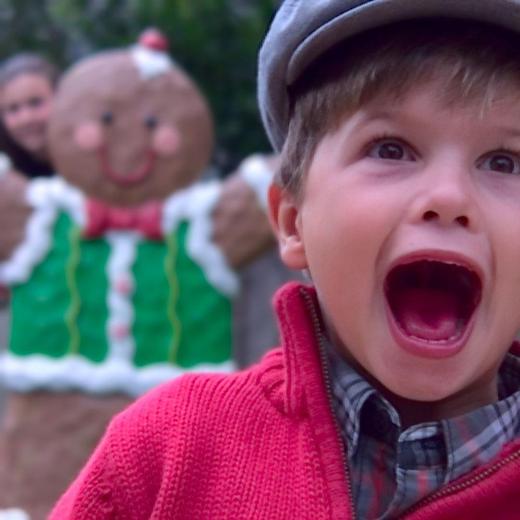}   & 
    \includegraphics[width=.24\linewidth]{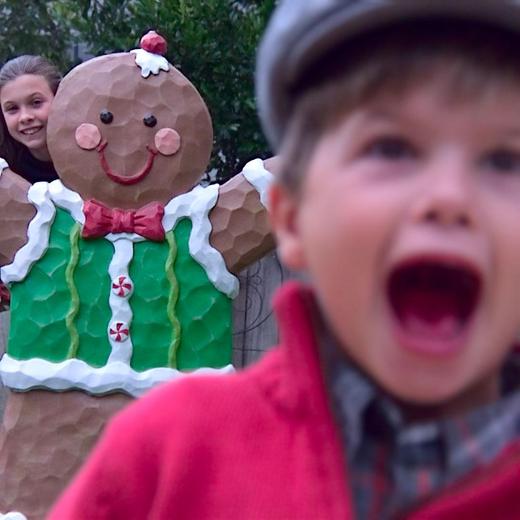}   & 
    \includegraphics[width=.24\linewidth]{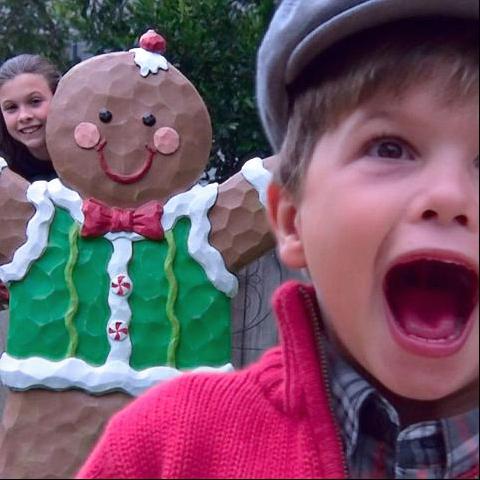}&
    \includegraphics[width=.24\linewidth]{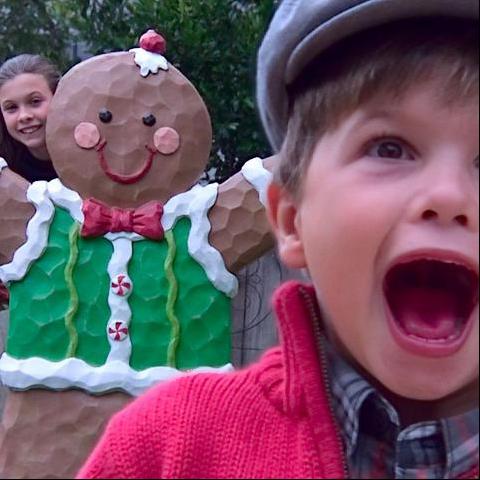}\\
    
    \includegraphics[width=.24\linewidth]{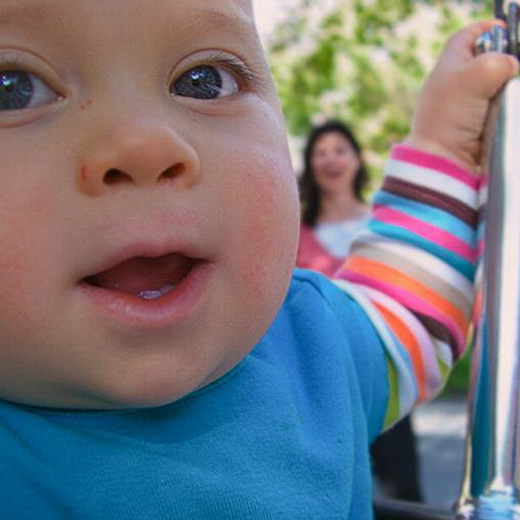}    &
    \includegraphics[width=.24\linewidth]{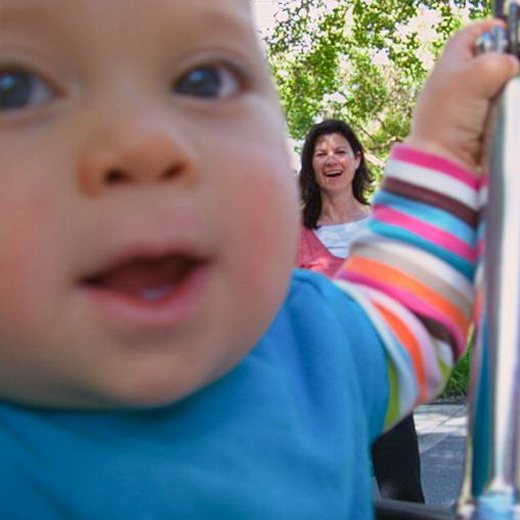}    &
    \includegraphics[width=.24\linewidth]{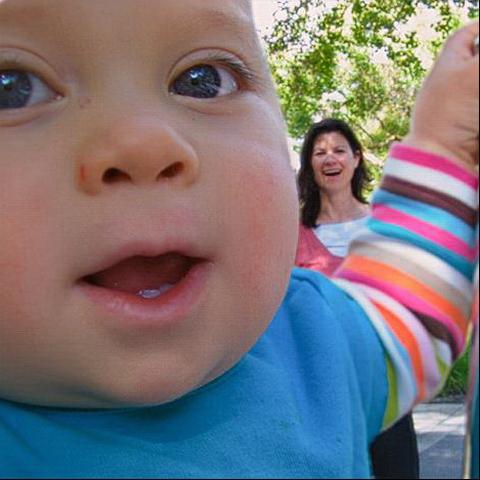}&
    \includegraphics[width=.24\linewidth]{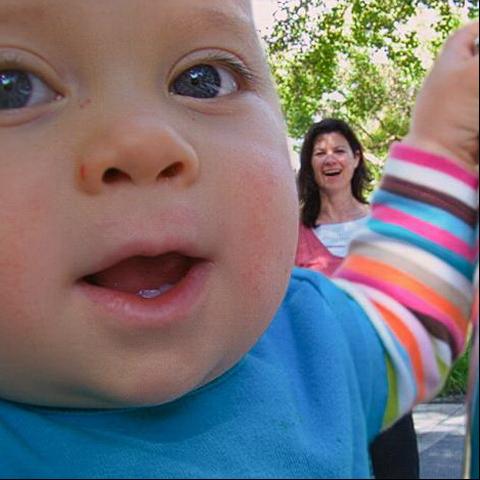}\\
    
    \end{tabular}
    
    \caption{Our fusion results over the Lytro 2 source real dataset.}
    \label{fig:res1}
\end{figure}

\begin{table*}[h!]
\caption{Mean and standard deviation of the objective assessment over the Lytro multi-focus two sources dataset.}
\label{tab:real_metrics}
\scriptsize
\resizebox{\linewidth}{!}{%
\begin{tabular}{rccccccc}
Metrics    &CNN \cite{liu2017multi} &  DCT+CV \cite{haghighat2011multi} & DCT \cite{haghighat2010real} & GFF\cite{li2013image1} & IM \cite{li2013image} & HF-Reg (Ours)             & HF-Seg (Ours) \\ \hline
$Q_{MI}$   &$1.1467 \pm 0.1107$     & $0.8476 \pm 0.1419$               & $0.8347 \pm 0.1403$          & $1.0932 \pm 0.1209$    & $1.1376 \pm 0.1045$   & $1.1538 \pm 0.0865$       & $\mathbf{1.1758 \pm 0.0968}$ \\
$Q_{TE}$   &$0.3994 \pm 0.0299$     & $0.3702 \pm 0.0380$               & $0.3656 \pm 0.0381$          & $0.3969 \pm 0.0320$    & $0.3961 \pm 0.0287$   & $0.3984 \pm 0.0268$       & $\mathbf{0.4020 \pm 0.0286}$ \\
$Q_{NCIE}$ &$0.8425 \pm 0.0080$     & $0.8259 \pm 0.0081$               & $0.8251 \pm 0.0077$          & $0.8390 \pm 0.0081$    & $0.8420 \pm 0.0078$   & $0.8423 \pm 0.0066$       & $\mathbf{0.8443 \pm 0.0076}$ \\
$Q_{G}$    &$\mathbf{0.7234 \pm 0.0280}$& $0.6939 \pm 0.0328$           & $0.6853 \pm 0.0353$          & $0.7182 \pm 0.0307$    & $0.7159 \pm 0.0301$   & $0.6636 \pm 0.0420$       & $0.7096 \pm 0.0315$ \\
$Q_{P}$    &$\mathbf{0.8488 \pm 0.0395}$& $0.8140 \pm 0.0490$           & $0.7633 \pm 0.0658$          & $0.8465 \pm 0.0395$    & $0.8205 \pm 0.0472$   & $0.8004 \pm 0.0432$       & $0.8387 \pm 0.0408$ \\
$Q_{S}$    &$0.9466 \pm 0.0124$     & $0.9377 \pm 0.0143$               & $0.9367 \pm 0.0147$          & $\mathbf{0.9467 \pm 0.0123}$& $0.9419 \pm 0.0131$& $0.9418 \pm 0.0131$     & $0.9447 \pm 0.0130$ \\
$Q_{CB}$   &$\mathbf{0.8058 \pm 0.0381}$& $0.7230 \pm 0.0395$           & $0.7030 \pm 0.0466$          & $0.7929 \pm 0.0400$    & $0.7922 \pm 0.0408$   & $0.7550 \pm 0.0477$       & $0.7898 \pm 0.0439$ \\
\end{tabular}
}
\end{table*}
The Lytro two sources dataset was used to evaluate our methods over real multi-focus images pairs. This dataset has 20 pairs of multi-focused images captured with the Lytro camera that uses the Light-field technology, allowing to expand the depth of field after the image was taken. Because the all-in-focus ground truth is not available, only the objective assessment metrics were used in this experiment. Table \ref{tab:real_metrics} shows that our method has a higher mean and lower variance in the first three metrics. 
%The comparison with and without Near post-processing shows that a better objective assessment is obtained with Near while no additional computational complexity is added. 
For all metrics there was not a statistically significant difference in the values of our proposal with CNN and GFF approaches. Some examples of the obtained all-in-focus images with our networks are shown in Fig. \ref{fig:res1}.

\begin{figure}[tb!]
    \footnotesize
    \setlength{\tabcolsep}{1pt}
    \begin{tabular}{ccc}
    Source A& Source B& CNN\\
    \includegraphics[width=.32\linewidth]{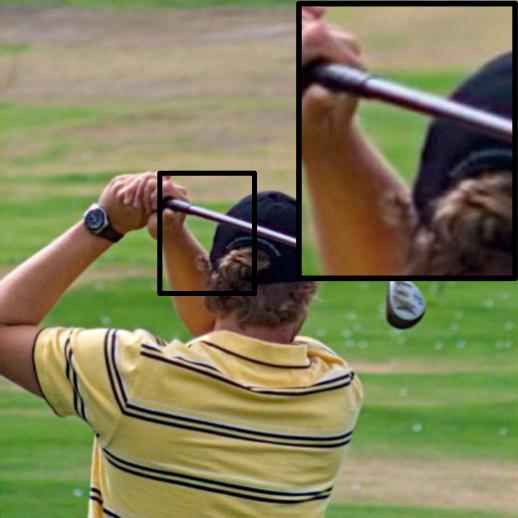}&    
    \includegraphics[width=.32\linewidth]{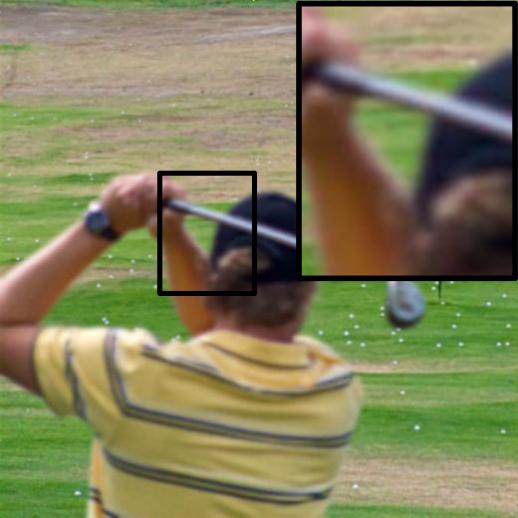}&    
    \includegraphics[width=.32\linewidth]{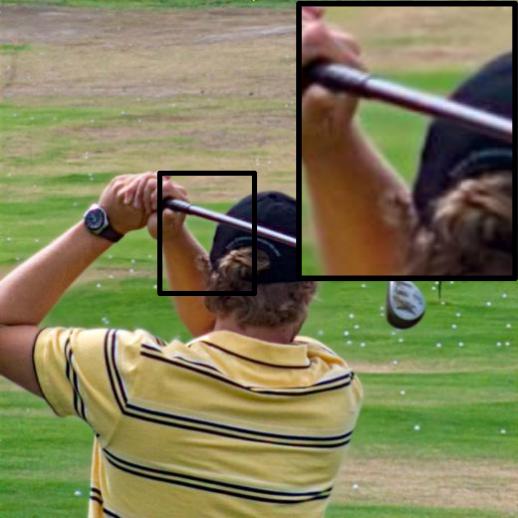}\\ 
    DCT+CV& DCT& GFF\\
    \includegraphics[width=.32\linewidth]{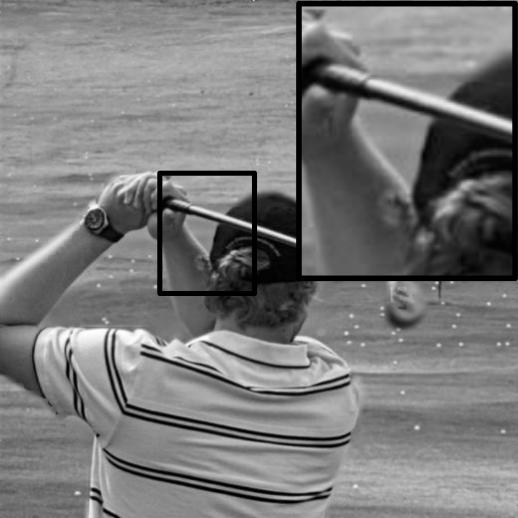}&
    \includegraphics[width=.32\linewidth]{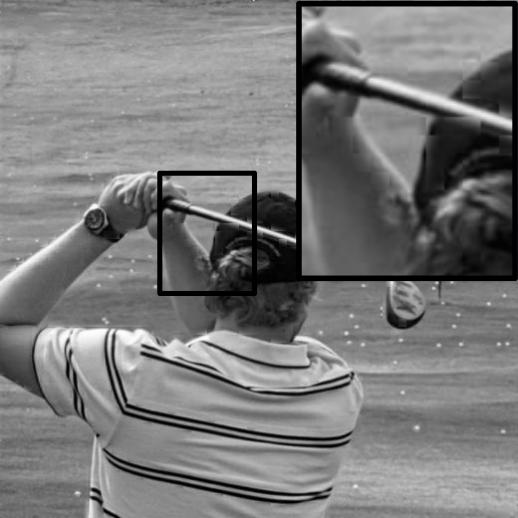}&
    \includegraphics[width=.32\linewidth]{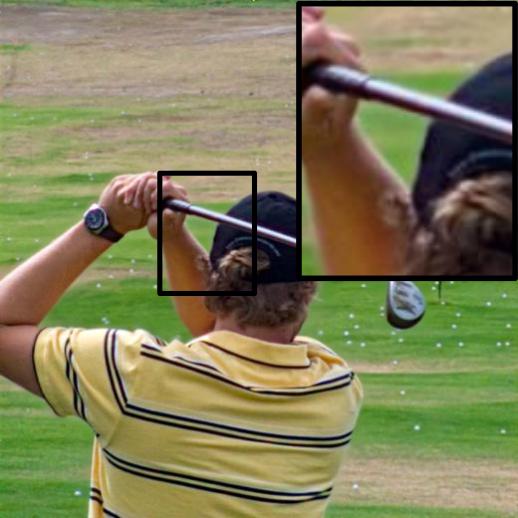}\\
    IM& HF-Reg& HF-Seg \\
    \includegraphics[width=.32\linewidth]{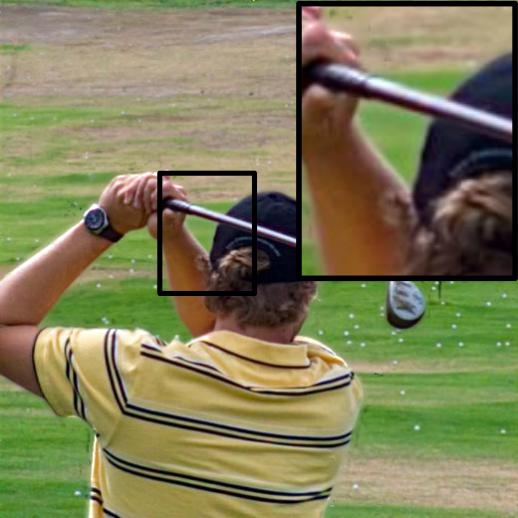}&
    \includegraphics[width=.32\linewidth]{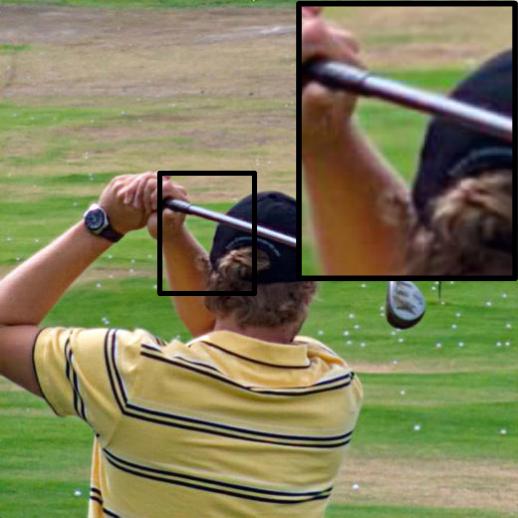}&
    \includegraphics[width=.32\linewidth]{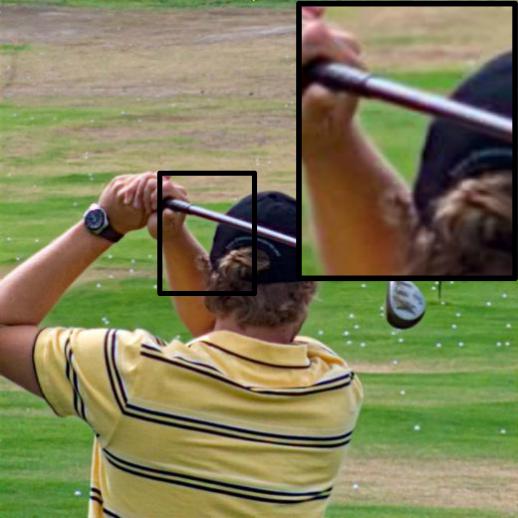}\\
    
    \end{tabular}
    
    \caption{Fusion results comparison with different literature methods and our proposal over "golf image" of the Lytro 2 dataset.}
    \label{fig:realores2}
\end{figure}

An advantage of our proposal is that we do not apply any further morphological operation in the post-processing step. The problem with this kind of operations is that the size and shape of the structural elements restrict the solution space. An example of this is shown in Fig. \ref{fig:realores2} for the "golf" image of the Lytro dataset. A visually comparable result is obtained with CNN, GFF, HF-Reg, and HF-Seg. 
%However, a closer inspection to the error image respect to Source A (Fig. \ref{fig:realres2}) reveals that the consistency verifications steps in CNN and GFF cause the wrong fusion in the zoomed gap region. 
However, a closer inspection into the marked area reveals that, contrary to our proposal, the consistency verification steps in CNN and GFF causes the wrong fusion in the gap region. 
%Also, in the figure is observed a dilated man shape for both methods making the fusion around the border blurred than in our proposal. 
%On the other hand our HF-Reg method showed a noisier regression in the high contrast region of the golf stick. We believe that further training of the HF-Reg network may solve this issue.

%\begin{figure*}[tb!]
%    \footnotesize
%    \setlength{\tabcolsep}{1pt}
%    \begin{tabular}{ccc}
%    CNN& DCT+CV& DCT\\
%    \includegraphics[width=.32\linewidth]{images/cnngolf.jpg}&    
%    \includegraphics[width=.32\linewidth]{images/dct1golf.jpg}&    
%    \includegraphics[width=.32\linewidth]{images/dct2golf.jpg}\\ 
%    GFF& IM& HF-Reg\\
%    \includegraphics[width=.32\linewidth]{images/gffgolf.jpg}&
%    \includegraphics[width=.32\linewidth]{images/ifmgolf.jpg}&
%    \includegraphics[width=.32\linewidth]{images/reggolf.jpg}\\
%    & HF-Seg& \\
%    &
%    \includegraphics[width=.32\linewidth]{images/seggolf.jpg}&
%    \\
%    
%    \end{tabular}
%    
%    \caption{Difference between Source A and every fusion result in Fig. \ref{fig:realores2}.}
%    \label{fig:realres2}
%\end{figure*}

\subsection{Three Sources real dataset}

\begin{figure}[tb!]
    \footnotesize
    \setlength{\tabcolsep}{1pt}
    \begin{tabular}{cccc}
    
    \multicolumn{4}{c}{Source A}\\
    \includegraphics[width=.24\linewidth]{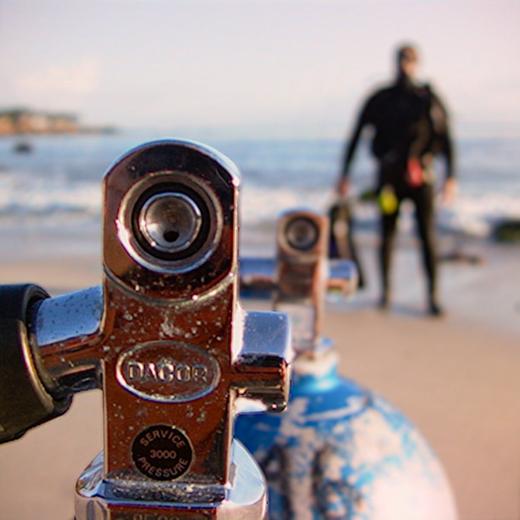}&    
    \includegraphics[width=.24\linewidth]{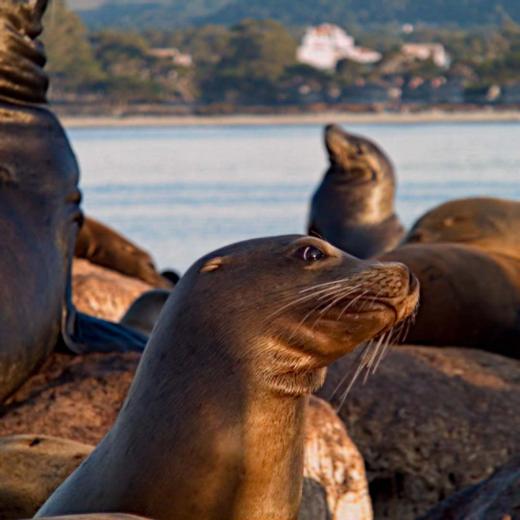}&  
    \includegraphics[width=.24\linewidth]{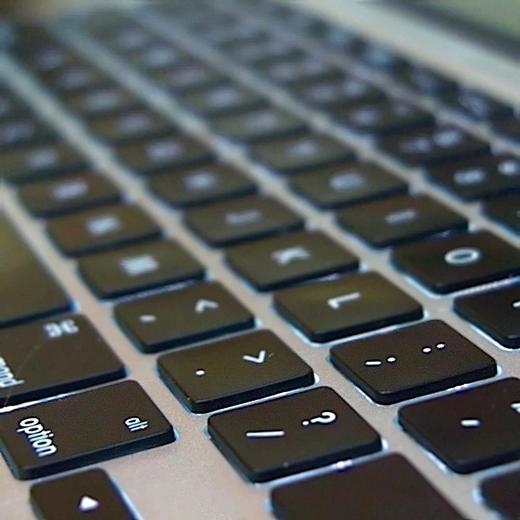}&    
    \includegraphics[width=.24\linewidth]{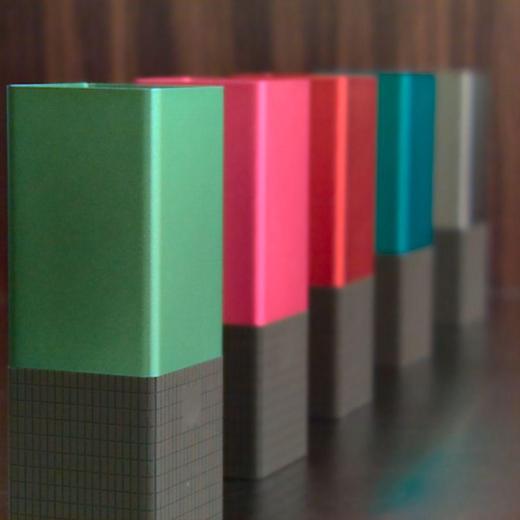}\\  
    
    \multicolumn{4}{c}{Source B}\\
    \includegraphics[width=.24\linewidth]{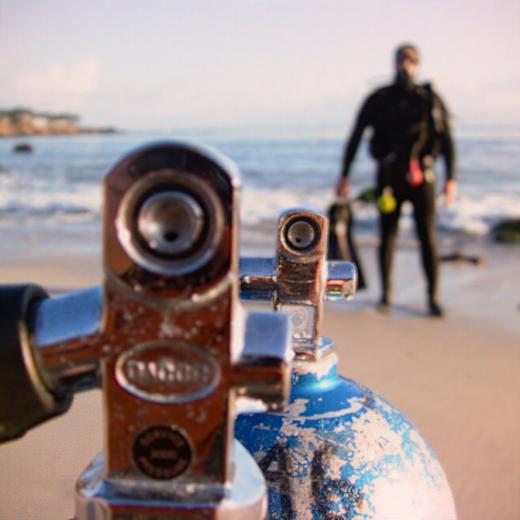}&    
    \includegraphics[width=.24\linewidth]{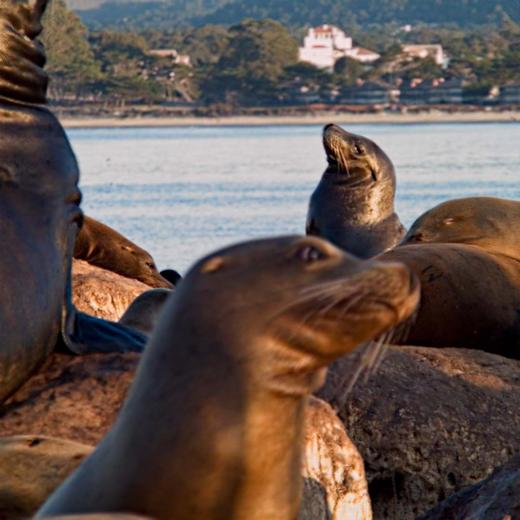}&    
    \includegraphics[width=.24\linewidth]{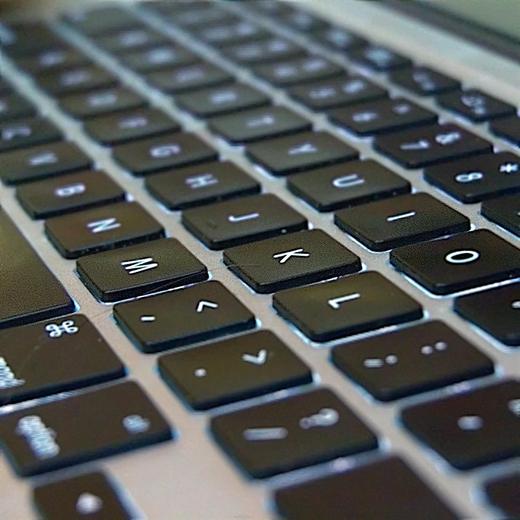}& 
    \includegraphics[width=.24\linewidth]{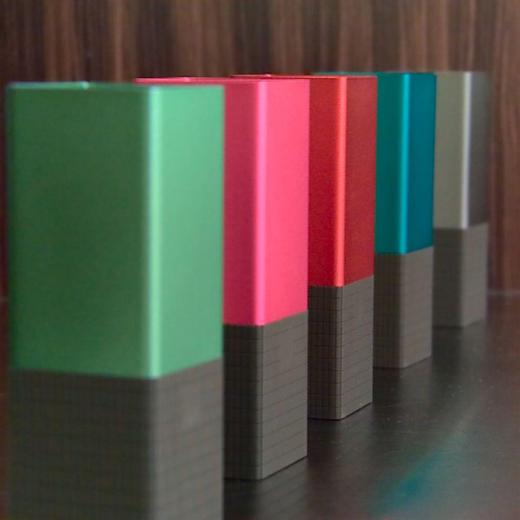}\\ 
    
    \multicolumn{4}{c}{Source C}\\
    \includegraphics[width=.24\linewidth]{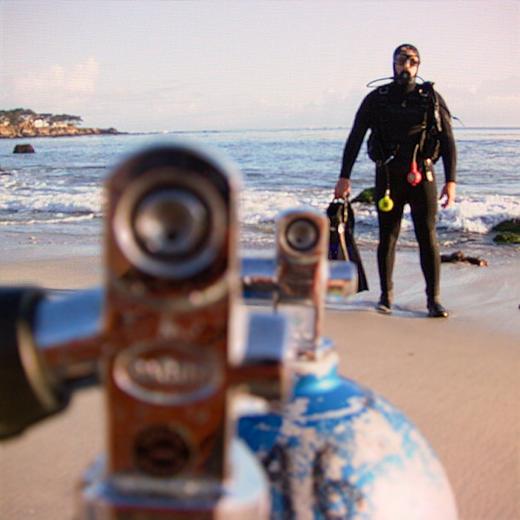}&    
    \includegraphics[width=.24\linewidth]{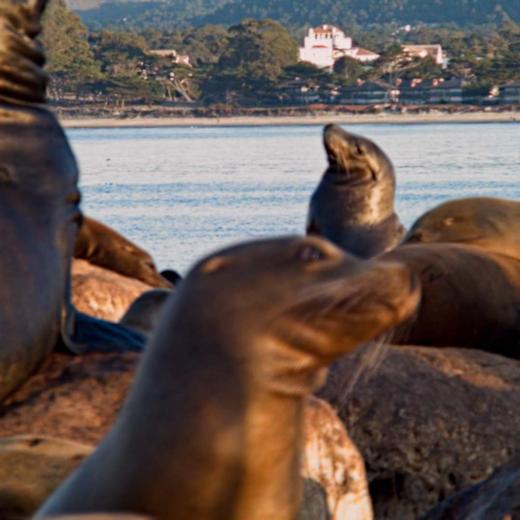}&  
    \includegraphics[width=.24\linewidth]{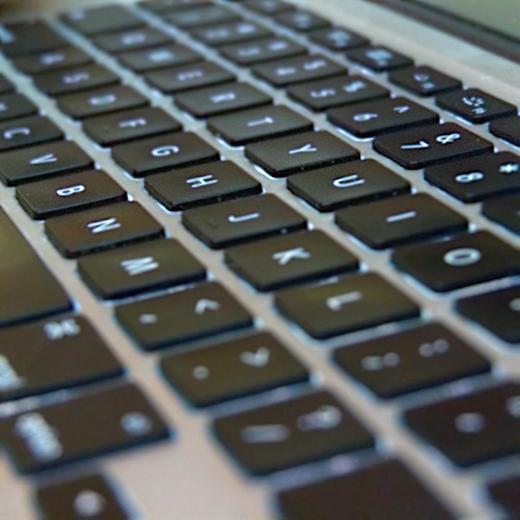}&    
    \includegraphics[width=.24\linewidth]{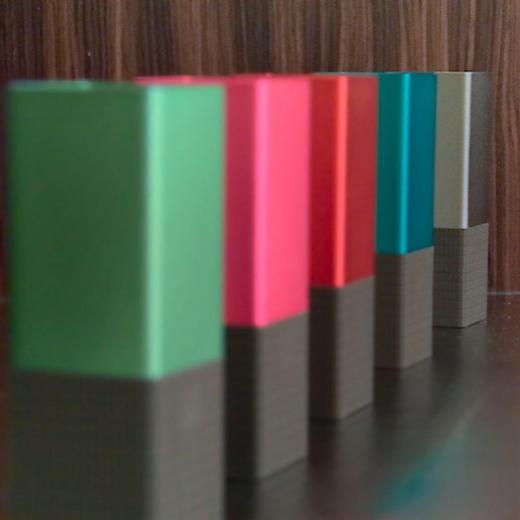}\\    
    
    \multicolumn{4}{c}{HF-Reg (Ours)}\\
    \includegraphics[width=.24\linewidth]{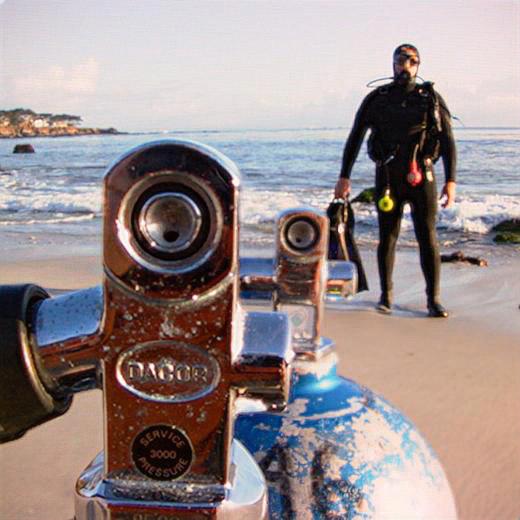}&
    \includegraphics[width=.24\linewidth]{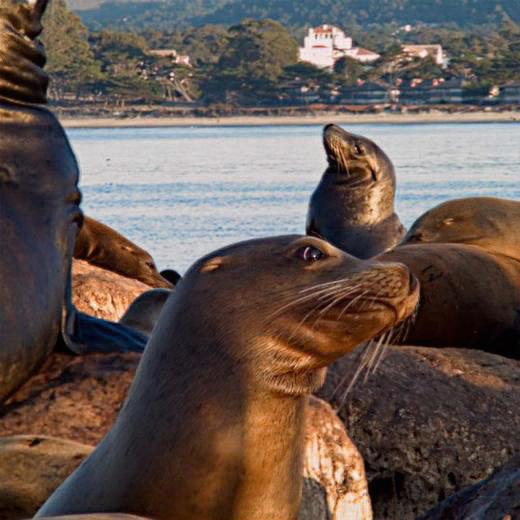}&
    \includegraphics[width=.24\linewidth]{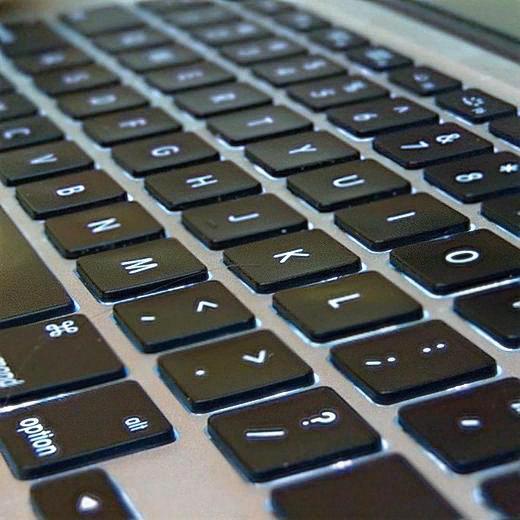}&
    \includegraphics[width=.24\linewidth]{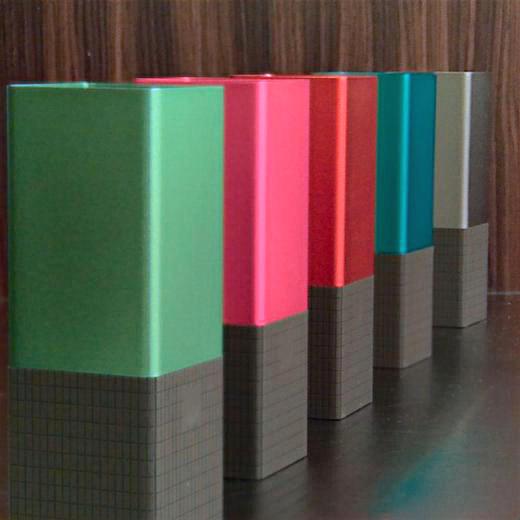}\\
    
    \multicolumn{4}{c}{HF-Seg (Ours)}\\
    \includegraphics[width=.24\linewidth]{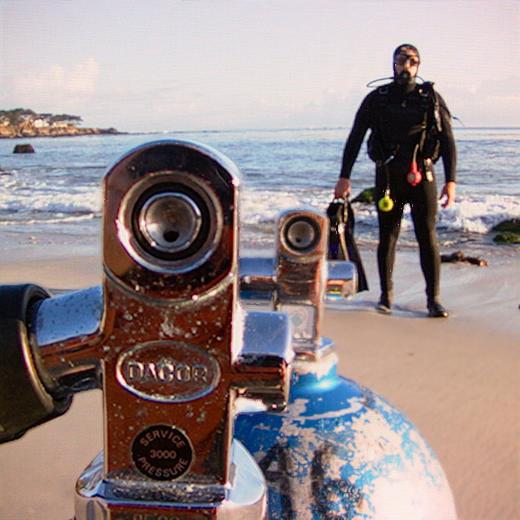}&
    \includegraphics[width=.24\linewidth]{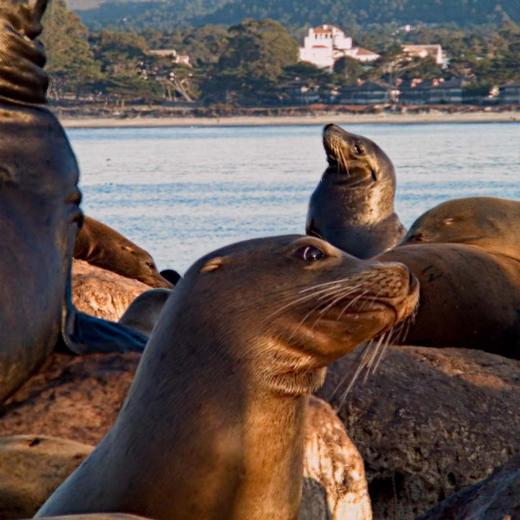}&  
    \includegraphics[width=.24\linewidth]{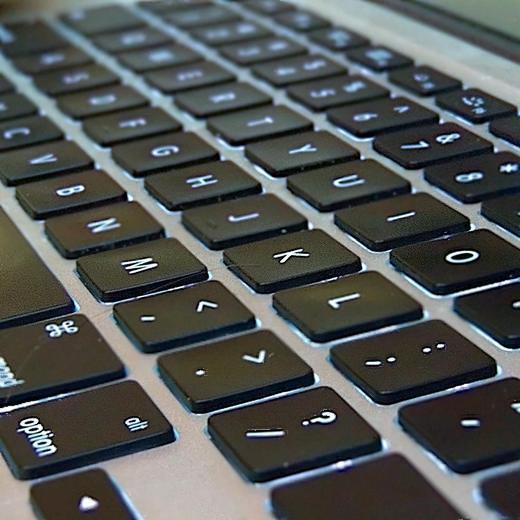}&
    \includegraphics[width=.24\linewidth]{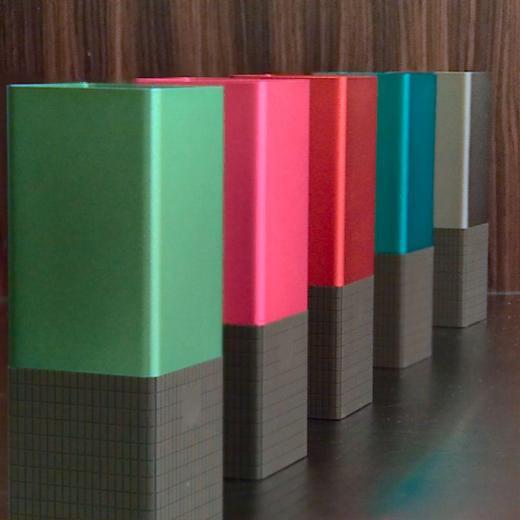}\\
    
    \end{tabular}
    
    \caption{Fusion results over the Lytro 3 sources real dataset.}
    \label{fig:res2}
\end{figure}

To show the performance of our method with multiple sources, we used the Lytro 3 sources real dataset. The dataset has four triplets of multi-focused images also captured with the Lytro camera. The $3$-functional power of fusion functions was computed in each case. Because the objective assessment metrics are defined for two sources, our evaluation was visual. As can be seen in Fig. \ref{fig:res2} our method can correctly obtain an all-in-focus image. Here, a better reconstruction is obtained with the HF-Seg network for the keyboard triplet fusion. This result is obtained because the accumulation of errors during the fusion is worst when a regression is done.

\subsection{Execution Time}
\begin{table}[b!]
\caption{Execution time for each mfif method with three different image size. Time unit is second.}
\label{tab:time}
\scriptsize
\resizebox{\linewidth}{!}{%
\begin{tabular}{rccc}
Method & $520 \times 520$ 	& $260 \times 260$ & $130 \times 130$   \\\hline
 CNN GPU (extracted from \cite{liu2017multi})& 0.7800 & - & -\\
 CNN slight GPU (extracted from \cite{liu2017multi})  & 0.3300 & - & -\\
%CNN (Matlab) & 94.7100 & 33.8200 & 12.6821 \\
IM	   & 2.7095 & 0.8112  & 0.4870\\
DCT+CV & 0.7648 & 0.2982  & 0.1949\\
GFF	   & 0.1280 & 0.0373  & 0.0149\\
HF-Seg  & \textbf{0.0026} & \textbf{0.0023}  & \textbf{0.0022}\\
HF-Reg & \textbf{0.0023} & \textbf{0.0022}  & \textbf{0.0021}\\
\end{tabular}
}
\end{table}

\begin{figure*}[t!]
    \footnotesize
    \setlength{\tabcolsep}{1pt}
    \begin{tabular}{ccc}
    Noisy Source A& Noisy Source B& HF-Reg\\
    \includegraphics[width=.32\linewidth]{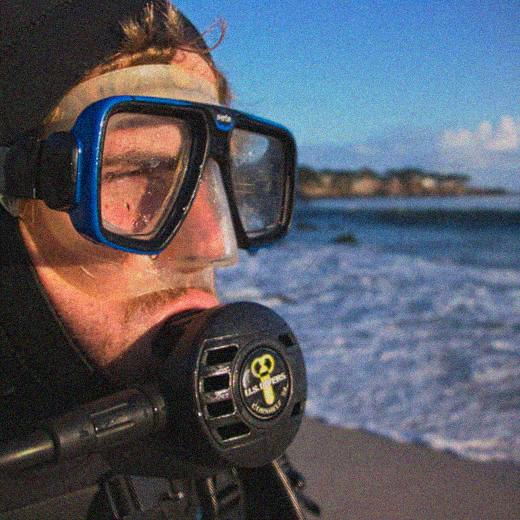}&    
    \includegraphics[width=.32\linewidth]{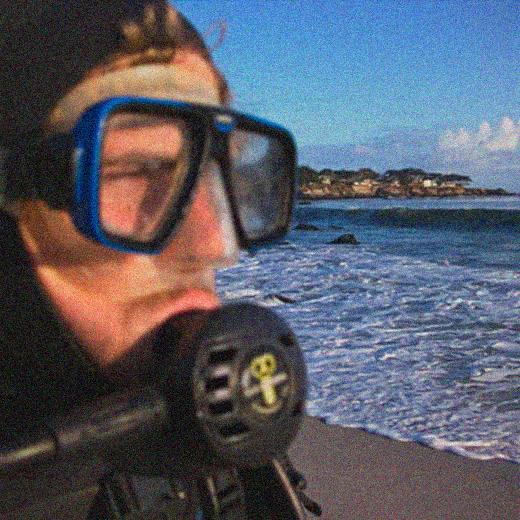}&    
    \includegraphics[width=.32\linewidth]{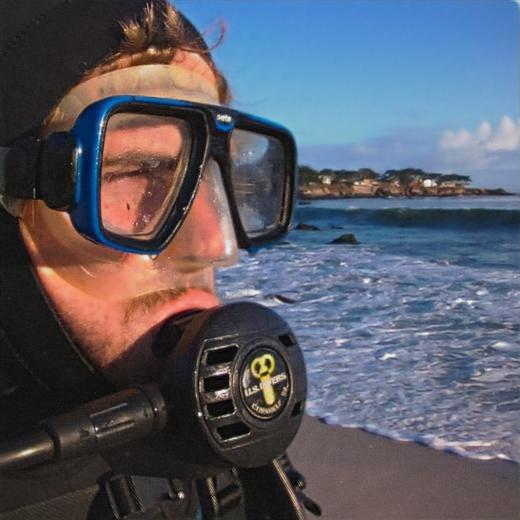}\\ 
    
    \end{tabular}
    
    \caption{Noisy source multi-focus image fusion.}
    \label{fig:filtered}
\end{figure*}

To test the execution time, we use the original implementations proposed by the authors of the tested methods. We are aware that exist a faster implementation of the CNN method than the one in Matlab, so for a fair evaluation, we included the time reported by the authors in their paper \cite{liu2017multi}. All methods were tested on the same computer with an Intel(R) Core(TM) i7-6800K 3.40 GHz CPU and 64GB RAM. Our approaches use a GPU GeForce GTX 1070 with PyTorch deep learning framework. We do not consider the time to load the data for any method. The synthetic multi-focus image dataset with 100 pairs was used for the experiment. Three different image sizes $520 \times 520$, $260 \times 260$ and $130 \times 130$ were tested. Table \ref{tab:time} shows the average execution time over the 100 images pairs. As can be seen, our fusion approaches have high computational efficiency when compared with the other methods. This outstanding computation makes our method good for near real-time applications where the multi-focus fusion is required. As shown in the previous experiments, this high efficiency does not decrease the performance, that is comparable or superior in most situations to the state-of-the-art.

\subsection{Applications of our HF-Reg}

Although our HF-Reg network does not outperform the HF-Seg approach according to the metrics, its idea is more straightforward, end-to-end, more general, and powerful that the other approaches revised. Because the regression problem is more complicated than the classification one, we believe that more training is needed for obtaining better solutions. The referred network can regress an image that does not need to be composed of pixels of the source, obtaining an improved filtered version. An example of this can be observed in Fig. \ref{fig:filtered} for an HF-Reg network trained during 500 epochs for the multi-focus fusion of noisy inputs. This kind of filtering and fusion was achieved by only applying Gaussian noise with a variable variance to the sources, and applying the previously described training protocol with NPS6.

%% file: images/loss.pdf_tex
%% Accompanies image file 'loss.pdf' (pdf, eps, ps)
\begingroup%
  \makeatletter%
  \providecommand\color[2][]{%
    \errmessage{(Inkscape) Color is used for the text in Inkscape, but the package 'color.sty' is not loaded}%
    \renewcommand\color[2][]{}%
  }%
  \providecommand\transparent[1]{%
    \errmessage{(Inkscape) Transparency is used (non-zero) for the text in Inkscape, but the package 'transparent.sty' is not loaded}%
    \renewcommand\transparent[1]{}%
  }%
  \providecommand\rotatebox[2]{#2}%
  \newcommand*\fsize{\dimexpr\f@size pt\relax}%
  \newcommand*\lineheight[1]{\fontsize{\fsize}{#1\fsize}\selectfont}%
  \ifx\svgwidth\undefined%
    \setlength{\unitlength}{373.01154848bp}%
    \ifx\svgscale\undefined%
      \relax%
    \else%
      \setlength{\unitlength}{\unitlength * \real{\svgscale}}%
    \fi%
  \else%
    \setlength{\unitlength}{\svgwidth}%
  \fi%
  \global\let\svgwidth\undefined%
  \global\let\svgscale\undefined%
  \makeatother%
  \begin{picture}(1,0.77815664)%
    \lineheight{1}%
    \setlength\tabcolsep{0pt}%
    \put(0,0){\includegraphics[width=\unitlength,page=1]{images/loss.pdf}}%
    \put(0.09493768,0.0455289){\color[rgb]{0.14901961,0.14901961,0.14901961}\makebox(0,0)[lt]{\lineheight{1.25}\smash{\scriptsize\begin{tabular}[t]{l}0\end{tabular}}}}%
    \put(0.25208797,0.0455289){\color[rgb]{0.14901961,0.14901961,0.14901961}\makebox(0,0)[lt]{\lineheight{1.25}\smash{\scriptsize\begin{tabular}[t]{l}200\end{tabular}}}}%
    \put(0.42653752,0.0455289){\color[rgb]{0.14901961,0.14901961,0.14901961}\makebox(0,0)[lt]{\lineheight{1.25}\smash{\scriptsize\begin{tabular}[t]{l}400\end{tabular}}}}%
    \put(0.60098702,0.0455289){\color[rgb]{0.14901961,0.14901961,0.14901961}\makebox(0,0)[lt]{\lineheight{1.25}\smash{\scriptsize\begin{tabular}[t]{l}600\end{tabular}}}}%
    \put(0.77543662,0.0455289){\color[rgb]{0.14901961,0.14901961,0.14901961}\makebox(0,0)[lt]{\lineheight{1.25}\smash{\scriptsize\begin{tabular}[t]{l}800\end{tabular}}}}%
    \put(0.94169179,0.0455289){\color[rgb]{0.14901961,0.14901961,0.14901961}\makebox(0,0)[lt]{\lineheight{1.25}\smash{\scriptsize\begin{tabular}[t]{l}1000\end{tabular}}}}%
    \put(0.48735854,0.00642858){\color[rgb]{0.14901961,0.14901961,0.14901961}\makebox(0,0)[lt]{\lineheight{1.25}\smash{\scriptsize\begin{tabular}[t]{l}Epochs\end{tabular}}}}%
    \put(0,0){\includegraphics[width=\unitlength,page=2]{images/loss.pdf}}%
    \put(0.04051073,0.07264136){\color[rgb]{0.14901961,0.14901961,0.14901961}\makebox(0,0)[lt]{\lineheight{1.25}\smash{\scriptsize\begin{tabular}[t]{l}-5.5\end{tabular}}}}%
    \put(0.06600441,0.14129213){\color[rgb]{0.14901961,0.14901961,0.14901961}\makebox(0,0)[lt]{\lineheight{1.25}\smash{\scriptsize\begin{tabular}[t]{l}-5\end{tabular}}}}%
    \put(0.04051073,0.2099429){\color[rgb]{0.14901961,0.14901961,0.14901961}\makebox(0,0)[lt]{\lineheight{1.25}\smash{\scriptsize\begin{tabular}[t]{l}-4.5\end{tabular}}}}%
    \put(0.06600441,0.27859379){\color[rgb]{0.14901961,0.14901961,0.14901961}\makebox(0,0)[lt]{\lineheight{1.25}\smash{\scriptsize\begin{tabular}[t]{l}-4\end{tabular}}}}%
    \put(0.04051073,0.34724456){\color[rgb]{0.14901961,0.14901961,0.14901961}\makebox(0,0)[lt]{\lineheight{1.25}\smash{\scriptsize\begin{tabular}[t]{l}-3.5\end{tabular}}}}%
    \put(0.06600441,0.41589533){\color[rgb]{0.14901961,0.14901961,0.14901961}\makebox(0,0)[lt]{\lineheight{1.25}\smash{\scriptsize\begin{tabular}[t]{l}-3\end{tabular}}}}%
    \put(0.04051073,0.48454622){\color[rgb]{0.14901961,0.14901961,0.14901961}\makebox(0,0)[lt]{\lineheight{1.25}\smash{\scriptsize\begin{tabular}[t]{l}-2.5\end{tabular}}}}%
    \put(0.06600441,0.55319699){\color[rgb]{0.14901961,0.14901961,0.14901961}\makebox(0,0)[lt]{\lineheight{1.25}\smash{\scriptsize\begin{tabular}[t]{l}-2\end{tabular}}}}%
    \put(0.04051073,0.62184782){\color[rgb]{0.14901961,0.14901961,0.14901961}\makebox(0,0)[lt]{\lineheight{1.25}\smash{\scriptsize\begin{tabular}[t]{l}-1.5\end{tabular}}}}%
    \put(0.06600441,0.6904987){\color[rgb]{0.14901961,0.14901961,0.14901961}\makebox(0,0)[lt]{\lineheight{1.25}\smash{\scriptsize\begin{tabular}[t]{l}-1\end{tabular}}}}%
    \put(0.04051073,0.75914947){\color[rgb]{0.14901961,0.14901961,0.14901961}\makebox(0,0)[lt]{\lineheight{1.25}\smash{\scriptsize\begin{tabular}[t]{l}-0.5\end{tabular}}}}%
    \put(0,0){\includegraphics[width=\unitlength,page=3]{images/loss.pdf}}%
    \put(0.88355182,0.71908799){\color[rgb]{0,0,0}\makebox(0,0)[lt]{\lineheight{1.25}\smash{\scriptsize\begin{tabular}[t]{l}NPS6\end{tabular}}}}%
    \put(0,0){\includegraphics[width=\unitlength,page=4]{images/loss.pdf}}%
    \put(0.88355182,0.68722083){\color[rgb]{0,0,0}\makebox(0,0)[lt]{\lineheight{1.25}\smash{\scriptsize\begin{tabular}[t]{l}L1\end{tabular}}}}%
    \put(0,0){\includegraphics[width=\unitlength,page=5]{images/loss.pdf}}%
    \put(0.88355182,0.65535379){\color[rgb]{0,0,0}\makebox(0,0)[lt]{\lineheight{1.25}\smash{\scriptsize\begin{tabular}[t]{l}MSE\end{tabular}}}}%
    \put(0,0){\includegraphics[width=\unitlength,page=6]{images/loss.pdf}}%
  \end{picture}%
\endgroup%

%% file: images/syn_ssim.pdf_tex
%% Accompanies image file 'syn_ssim.pdf' (pdf, eps, ps)
\begingroup%
  \makeatletter%
  \providecommand\color[2][]{%
    \errmessage{(Inkscape) Color is used for the text in Inkscape, but the package 'color.sty' is not loaded}%
    \renewcommand\color[2][]{}%
  }%
  \providecommand\transparent[1]{%
    \errmessage{(Inkscape) Transparency is used (non-zero) for the text in Inkscape, but the package 'transparent.sty' is not loaded}%
    \renewcommand\transparent[1]{}%
  }%
  \providecommand\rotatebox[2]{#2}%
  \newcommand*\fsize{\dimexpr\f@size pt\relax}%
  \newcommand*\lineheight[1]{\fontsize{\fsize}{#1\fsize}\selectfont}%
  \ifx\svgwidth\undefined%
    \setlength{\unitlength}{367.2132721bp}%
    \ifx\svgscale\undefined%
      \relax%
    \else%
      \setlength{\unitlength}{\unitlength * \real{\svgscale}}%
    \fi%
  \else%
    \setlength{\unitlength}{\svgwidth}%
  \fi%
  \global\let\svgwidth\undefined%
  \global\let\svgscale\undefined%
  \makeatother%
  \begin{picture}(1,0.73530814)%
    \lineheight{1}%
    \setlength\tabcolsep{0pt}%
    \put(0,0){\includegraphics[width=\unitlength,page=1]{images/syn_ssim.pdf}}%
    \put(0.14819433,0.00049017){\color[rgb]{0.14901961,0.14901961,0.14901961}\makebox(0,0)[lt]{\lineheight{1.25}\smash{\scriptsize\begin{tabular}[t]{l}CNN\end{tabular}}}}%
    \put(0.26921948,0.00049017){\color[rgb]{0.14901961,0.14901961,0.14901961}\makebox(0,0)[lt]{\lineheight{1.25}\smash{\scriptsize\begin{tabular}[t]{l}DCT+CV\end{tabular}}}}%
    \put(0.39579383,0.00049017){\color[rgb]{0.14901961,0.14901961,0.14901961}\makebox(0,0)[lt]{\lineheight{1.25}\smash{\scriptsize\begin{tabular}[t]{l}DCT\end{tabular}}}}%
    \put(0.53439143,0.00049017){\color[rgb]{0.14901961,0.14901961,0.14901961}\makebox(0,0)[lt]{\lineheight{1.25}\smash{\scriptsize\begin{tabular}[t]{l}GFF\end{tabular}}}}%
    \put(0.65726636,0.00049017){\color[rgb]{0.14901961,0.14901961,0.14901961}\makebox(0,0)[lt]{\lineheight{1.25}\smash{\scriptsize\begin{tabular}[t]{l}IM\end{tabular}}}}%
    \put(0.77459207,0.00049017){\color[rgb]{0.14901961,0.14901961,0.14901961}\makebox(0,0)[lt]{\lineheight{1.25}\smash{\scriptsize\begin{tabular}[t]{l}HF-REG\end{tabular}}}}%
    \put(0.89931667,0.00049017){\color[rgb]{0.14901961,0.14901961,0.14901961}\makebox(0,0)[lt]{\lineheight{1.25}\smash{\scriptsize\begin{tabular}[t]{l}HF-SEG\end{tabular}}}}%
    \put(0,0){\includegraphics[width=\unitlength,page=2]{images/syn_ssim.pdf}}%
    \put(0.05459212,0.0452777){\color[rgb]{0.14901961,0.14901961,0.14901961}\makebox(0,0)[lt]{\lineheight{1.25}\smash{\scriptsize\begin{tabular}[t]{l}0.9\end{tabular}}}}%
    \put(0.03794455,0.11011838){\color[rgb]{0.14901961,0.14901961,0.14901961}\makebox(0,0)[lt]{\lineheight{1.25}\smash{\scriptsize\begin{tabular}[t]{l}0.91\end{tabular}}}}%
    \put(0.03794455,0.17495859){\color[rgb]{0.14901961,0.14901961,0.14901961}\makebox(0,0)[lt]{\lineheight{1.25}\smash{\scriptsize\begin{tabular}[t]{l}0.92\end{tabular}}}}%
    \put(0.03794455,0.23979891){\color[rgb]{0.14901961,0.14901961,0.14901961}\makebox(0,0)[lt]{\lineheight{1.25}\smash{\scriptsize\begin{tabular}[t]{l}0.93\end{tabular}}}}%
    \put(0.03794455,0.30463924){\color[rgb]{0.14901961,0.14901961,0.14901961}\makebox(0,0)[lt]{\lineheight{1.25}\smash{\scriptsize\begin{tabular}[t]{l}0.94\end{tabular}}}}%
    \put(0.03794455,0.36947944){\color[rgb]{0.14901961,0.14901961,0.14901961}\makebox(0,0)[lt]{\lineheight{1.25}\smash{\scriptsize\begin{tabular}[t]{l}0.95\end{tabular}}}}%
    \put(0.03794455,0.43431977){\color[rgb]{0.14901961,0.14901961,0.14901961}\makebox(0,0)[lt]{\lineheight{1.25}\smash{\scriptsize\begin{tabular}[t]{l}0.96\end{tabular}}}}%
    \put(0.03794455,0.49916045){\color[rgb]{0.14901961,0.14901961,0.14901961}\makebox(0,0)[lt]{\lineheight{1.25}\smash{\scriptsize\begin{tabular}[t]{l}0.97\end{tabular}}}}%
    \put(0.03794455,0.56400071){\color[rgb]{0.14901961,0.14901961,0.14901961}\makebox(0,0)[lt]{\lineheight{1.25}\smash{\scriptsize\begin{tabular}[t]{l}0.98\end{tabular}}}}%
    \put(0.03794455,0.62884104){\color[rgb]{0.14901961,0.14901961,0.14901961}\makebox(0,0)[lt]{\lineheight{1.25}\smash{\scriptsize\begin{tabular}[t]{l}0.99\end{tabular}}}}%
    \put(0.08048834,0.69368124){\color[rgb]{0.14901961,0.14901961,0.14901961}\makebox(0,0)[lt]{\lineheight{1.25}\smash{\scriptsize\begin{tabular}[t]{l}1\end{tabular}}}}%
    \put(0.02237603,0.34895917){\color[rgb]{0.14901961,0.14901961,0.14901961}\rotatebox{90.000003}{\makebox(0,0)[lt]{\scriptsize\lineheight{1.25}\smash{\begin{tabular}[t]{l}SSIM\end{tabular}}}}}%
    \put(0,0){\includegraphics[width=\unitlength,page=3]{images/syn_ssim.pdf}}%
  \end{picture}%
\endgroup%

%% file: images/syn_metrics.pdf_tex
%% Accompanies image file 'syn_metrics.pdf' (pdf, eps, ps)
\begingroup%
  \makeatletter%
  \providecommand\color[2][]{%
    \errmessage{(Inkscape) Color is used for the text in Inkscape, but the package 'color.sty' is not loaded}%
    \renewcommand\color[2][]{}%
  }%
  \providecommand\transparent[1]{%
    \errmessage{(Inkscape) Transparency is used (non-zero) for the text in Inkscape, but the package 'transparent.sty' is not loaded}%
    \renewcommand\transparent[1]{}%
  }%
  \providecommand\rotatebox[2]{#2}%
  \newcommand*\fsize{\dimexpr\f@size pt\relax}%
  \newcommand*\lineheight[1]{\fontsize{\fsize}{#1\fsize}\selectfont}%
  \ifx\svgwidth\undefined%
    \setlength{\unitlength}{350.2350678bp}%
    \ifx\svgscale\undefined%
      \relax%
    \else%
      \setlength{\unitlength}{\unitlength * \real{\svgscale}}%
    \fi%
  \else%
    \setlength{\unitlength}{\svgwidth}%
  \fi%
  \global\let\svgwidth\undefined%
  \global\let\svgscale\undefined%
  \makeatother%
  \begin{picture}(1,0.79162685)%
    \lineheight{1}%
    \setlength\tabcolsep{0pt}%
    \put(0,0){\includegraphics[width=\unitlength,page=1]{images/syn_metrics.pdf}}%
    \put(0.08658735,0.00359753){\color[rgb]{0.14901961,0.14901961,0.14901961}\makebox(0,0)[lt]{\lineheight{1.25}\smash{\scriptsize\begin{tabular}[t]{l}$Q_{MI}$\end{tabular}}}}%
    \put(0.20370451,0.00359753){\color[rgb]{0.14901961,0.14901961,0.14901961}\makebox(0,0)[lt]{\lineheight{1.25}\smash{\scriptsize\begin{tabular}[t]{l}$Q_{TE}$\end{tabular}}}}%
    \put(0.3051448,0.00359753){\color[rgb]{0.14901961,0.14901961,0.14901961}\makebox(0,0)[lt]{\lineheight{1.25}\smash{\scriptsize\begin{tabular}[t]{l}$Q_{NCIE}$\end{tabular}}}}%
    \put(0.44666597,0.00359753){\color[rgb]{0.14901961,0.14901961,0.14901961}\makebox(0,0)[lt]{\lineheight{1.25}\smash{\scriptsize\begin{tabular}[t]{l}$Q_G$\end{tabular}}}}%
    \put(0.56669218,0.00359753){\color[rgb]{0.14901961,0.14901961,0.14901961}\makebox(0,0)[lt]{\lineheight{1.25}\smash{\scriptsize\begin{tabular}[t]{l}$Q_P$\end{tabular}}}}%
    \put(0.68477897,0.00359753){\color[rgb]{0.14901961,0.14901961,0.14901961}\makebox(0,0)[lt]{\lineheight{1.25}\smash{\scriptsize\begin{tabular}[t]{l}$Q_S$\end{tabular}}}}%
    \put(0.77377486,0.00359753){\color[rgb]{0.14901961,0.14901961,0.14901961}\makebox(0,0)[lt]{\lineheight{1.25}\smash{\scriptsize\begin{tabular}[t]{l}$Q_{CB}$\end{tabular}}}}%
    \put(0.90640713,0.00359753){\color[rgb]{0.14901961,0.14901961,0.14901961}\makebox(0,0)[lt]{\lineheight{1.25}\smash{\scriptsize\begin{tabular}[t]{l}SSIM\end{tabular}}}}%
    \put(0,0){\includegraphics[width=\unitlength,page=2]{images/syn_metrics.pdf}}%
    \put(0.02592384,0.032473){\color[rgb]{0.14901961,0.14901961,0.14901961}\makebox(0,0)[lt]{\lineheight{1.25}\smash{\scriptsize\begin{tabular}[t]{l}0\end{tabular}}}}%
    \put(-0.00122773,0.15562473){\color[rgb]{0.14901961,0.14901961,0.14901961}\makebox(0,0)[lt]{\lineheight{1.25}\smash{\scriptsize\begin{tabular}[t]{l}0.2\end{tabular}}}}%
    \put(-0.00122773,0.27877654){\color[rgb]{0.14901961,0.14901961,0.14901961}\makebox(0,0)[lt]{\lineheight{1.25}\smash{\scriptsize\begin{tabular}[t]{l}0.4\end{tabular}}}}%
    \put(-0.00122773,0.40192827){\color[rgb]{0.14901961,0.14901961,0.14901961}\makebox(0,0)[lt]{\lineheight{1.25}\smash{\scriptsize\begin{tabular}[t]{l}0.6\end{tabular}}}}%
    \put(-0.00122773,0.52508007){\color[rgb]{0.14901961,0.14901961,0.14901961}\makebox(0,0)[lt]{\lineheight{1.25}\smash{\scriptsize\begin{tabular}[t]{l}0.8\end{tabular}}}}%
    \put(0.02592384,0.64823188){\color[rgb]{0.14901961,0.14901961,0.14901961}\makebox(0,0)[lt]{\lineheight{1.25}\smash{\scriptsize\begin{tabular}[t]{l}1\end{tabular}}}}%
    \put(0,0){\includegraphics[width=\unitlength,page=3]{images/syn_metrics.pdf}}%
    \put(0.49714662,0.73467362){\color[rgb]{0,0,0}\makebox(0,0)[lt]{\lineheight{1.25}\smash{\scriptsize\begin{tabular}[t]{l}HF-Seg\end{tabular}}}}%
    \put(0,0){\includegraphics[width=\unitlength,page=4]{images/syn_metrics.pdf}}%
    \put(0.49714662,0.70096502){\color[rgb]{0,0,0}\makebox(0,0)[lt]{\lineheight{1.25}\smash{\scriptsize\begin{tabular}[t]{l}HF-Reg\end{tabular}}}}%
    \put(0,0){\includegraphics[width=\unitlength,page=5]{images/syn_metrics.pdf}}%
    \put(0.49714662,0.66725643){\color[rgb]{0,0,0}\makebox(0,0)[lt]{\lineheight{1.25}\smash{\scriptsize\begin{tabular}[t]{l}GFF\end{tabular}}}}%
    \put(0,0){\includegraphics[width=\unitlength,page=6]{images/syn_metrics.pdf}}%
    \put(0.49714662,0.63354784){\color[rgb]{0,0,0}\makebox(0,0)[lt]{\lineheight{1.25}\smash{\scriptsize\begin{tabular}[t]{l}CNN\end{tabular}}}}%
    \put(0,0){\includegraphics[width=\unitlength,page=7]{images/syn_metrics.pdf}}%
  \end{picture}%
\endgroup%

%% file: ms.bbl
\begin{thebibliography}{10}
\providecommand{\url}[1]{#1}
\csname url@samestyle\endcsname
\providecommand{\newblock}{\relax}
\providecommand{\bibinfo}[2]{#2}
\providecommand{\BIBentrySTDinterwordspacing}{\spaceskip=0pt\relax}
\providecommand{\BIBentryALTinterwordstretchfactor}{4}
\providecommand{\BIBentryALTinterwordspacing}{\spaceskip=\fontdimen2\font plus
\BIBentryALTinterwordstretchfactor\fontdimen3\font minus
  \fontdimen4\font\relax}
\providecommand{\BIBforeignlanguage}[2]{{%
\expandafter\ifx\csname l@#1\endcsname\relax
\typeout{** WARNING: IEEEtran.bst: No hyphenation pattern has been}%
\typeout{** loaded for the language `#1'. Using the pattern for}%
\typeout{** the default language instead.}%
\else
\language=\csname l@#1\endcsname
\fi
#2}}
\providecommand{\BIBdecl}{\relax}
\BIBdecl

\bibitem{tan2017automated}
J.~H. Tan, H.~Fujita, S.~Sivaprasad, S.~V. Bhandary, A.~K. Rao, K.~C. Chua, and
  U.~R. Acharya, ``Automated segmentation of exudates, haemorrhages,
  microaneurysms using single convolutional neural network,'' \emph{Information
  sciences}, vol. 420, pp. 66--76, 2017.

\bibitem{tang2018pixel}
H.~Tang, B.~Xiao, W.~Li, and G.~Wang, ``Pixel convolutional neural network for
  multi-focus image fusion,'' \emph{Information Sciences}, vol. 433, pp.
  125--141, 2018.

\bibitem{gangapure2015steerable}
V.~N. Gangapure, S.~Banerjee, and A.~S. Chowdhury, ``Steerable local frequency
  based multispectral multifocus image fusion,'' \emph{Information fusion},
  vol.~23, pp. 99--115, 2015.

\bibitem{kong2014adaptive}
W.~Kong, Y.~Lei, and H.~Zhao, ``Adaptive fusion method of visible light and
  infrared images based on non-subsampled shearlet transform and fast
  non-negative matrix factorization,'' \emph{Infrared Physics \& Technology},
  vol.~67, pp. 161--172, 2014.

\bibitem{yan2018unsupervised}
X.~Yan, S.~Z. Gilani, H.~Qin, and A.~Mian, ``Unsupervised deep multi-focus
  image fusion,'' \emph{arXiv preprint arXiv:1806.07272}, 2018.

\bibitem{liu2017multi}
Y.~Liu, X.~Chen, H.~Peng, and Z.~Wang, ``Multi-focus image fusion with a deep
  convolutional neural network,'' \emph{Information Fusion}, vol.~36, pp.
  191--207, 2017.

\bibitem{zagoruyko2015learning}
S.~Zagoruyko and N.~Komodakis, ``Learning to compare image patches via
  convolutional neural networks,'' in \emph{IEEE Conference on Computer Vision
  and Pattern Recognition (CVPR)}, 2015, pp. 4353--4361.

\bibitem{nejati2015multi}
M.~Nejati, S.~Samavi, and S.~Shirani, ``Multi-focus image fusion using
  dictionary-based sparse representation,'' \emph{Information Fusion}, vol.~25,
  pp. 72--84, 2015.

\bibitem{petrovic2004gradient}
V.~S. Petrovic and C.~S. Xydeas, ``Gradient-based multiresolution image
  fusion,'' \emph{IEEE Transactions on Image processing}, vol.~13, no.~2, pp.
  228--237, 2004.

\bibitem{lewis2007pixel}
J.~J. Lewis, R.~J. O’Callaghan, S.~G. Nikolov, D.~R. Bull, and
  N.~Canagarajah, ``Pixel-and region-based image fusion with complex
  wavelets,'' \emph{Information fusion}, vol.~8, no.~2, pp. 119--130, 2007.

\bibitem{zhang2009multifocus}
Q.~Zhang and B.-l. Guo, ``Multifocus image fusion using the nonsubsampled
  contourlet transform,'' \emph{Signal processing}, vol.~89, no.~7, pp.
  1334--1346, 2009.

\bibitem{haghighat2010real}
A.~Haghighat, A.~Aghagolzadeh, and H.~Seyedarabi, ``{Real-time fusion of
  multi-focus images for visual sensor networks},'' in \emph{Machine Vision and
  Image Processing (MVIP), 2010 6th Iranian}.\hskip 1em plus 0.5em minus
  0.4em\relax IEEE, 2010, pp. 1--6.

\bibitem{haghighat2011multi}
{A. Haghighat, A. Aghagolzadeh and H. Seyedarabi}, ``{Multi-focus image fusion
  for visual sensor networks in DCT domain},'' \emph{Computers and Electrical
  Engineering}, vol.~37, no.~5, pp. 789--797, 2011.

\bibitem{yang2010multifocus}
B.~Yang and S.~Li, ``Multifocus image fusion and restoration with sparse
  representation,'' \emph{IEEE Transactions on Instrumentation and
  Measurement}, vol.~59, no.~4, pp. 884--892, 2010.

\bibitem{li2013image}
S.~Li, X.~Kang, J.~Hu, and B.~Yang, ``Image matting for fusion of multi-focus
  images in dynamic scenes,'' \emph{Information Fusion}, vol.~14, no.~2, pp.
  147--162, 2013.

\bibitem{li2013image1}
S.~Li, X.~Kang, and J.~Hu, ``Image fusion with guided filtering,'' \emph{IEEE
  Transactions on Image Processing}, vol.~22, no.~7, pp. 2864--2875, 2013.

\bibitem{ronneberger2015u}
O.~Ronneberger, P.~Fischer, and T.~Brox, ``U-net: Convolutional networks for
  biomedical image segmentation,'' in \emph{International Conference on Medical
  image computing and computer-assisted intervention}.\hskip 1em plus 0.5em
  minus 0.4em\relax Springer, 2015, pp. 234--241.

\bibitem{lin2014microsoft}
T.-Y. Lin, M.~Maire, S.~Belongie, J.~Hays, P.~Perona, D.~Ramanan,
  P.~Doll{\'a}r, and C.~L. Zitnick, ``{Microsoft COCO: Common objects in
  context},'' in \emph{European Conference on Computer Vision (ECCV)}, 2014,
  pp. 740--755.

\bibitem{kingma2014adam}
D.~P. Kingma and J.~Ba, ``Adam: A method for stochastic optimization,''
  \emph{arXiv preprint arXiv:1412.6980}, 2014.

\bibitem{glorot2010understanding}
X.~Glorot and Y.~Bengio, ``Understanding the difficulty of training deep
  feedforward neural networks,'' in \emph{Proceedings of the thirteenth
  international conference on artificial intelligence and statistics}, 2010,
  pp. 249--256.

\bibitem{liu2012objective}
Z.~Liu, E.~Blasch, Z.~Xue, J.~Zhao, R.~Laganiere, and W.~Wu, ``Objective
  assessment of multiresolution image fusion algorithms for context enhancement
  in night vision: a comparative study,'' \emph{IEEE transactions on pattern
  analysis and machine intelligence}, vol.~34, no.~1, pp. 94--109, 2012.

\end{thebibliography}
